\documentclass[journal]{IEEEtran}
\ifCLASSINFOpdf
\else
\fi

\usepackage{amsmath,amsfonts,amssymb}
\usepackage{algpseudocode}

\usepackage[ruled,linesnumbered]{algorithm2e}  
\usepackage{array}
\usepackage{textcomp}
\usepackage{stfloats}
\usepackage{url}
\usepackage{verbatim}
\usepackage{graphicx}
\usepackage{caption}
\usepackage{subfigure}
\usepackage{cite}
\usepackage{multirow}
\usepackage{threeparttable}
\usepackage{hyperref}
\usepackage{multirow}
\usepackage{longtable}
\usepackage{xcolor}
\usepackage{tablefootnote}
\usepackage{enumitem}
\usepackage[T1]{fontenc}
\usepackage{textcomp}

\def\BibTeX{{\rm B\kern-.05em{\sc i\kern-.025em b}\kern-.08em
    T\kern-.1667em\lower.7ex\hbox{E}\kern-.125emX}}

\begin{document}
%
\title{TransGP: Task-Conditioned Transformer-Guided Genetic Programming for Multitask Dynamic Flexible Job Shop Scheduling}
%
%
%

\author{Meng~Xu, Jiao~Liu, Hua~Yu, and~Yew~Soon~Ong,~\IEEEmembership{Fellow,~IEEE}}


\markboth{Journal of \LaTeX\ Class Files,~Vol.~14, No.~8, August~2015}%
{Shell \MakeLowercase{\textit{et al.}}: Bare Demo of IEEEtran.cls for IEEE Journals}
%



\maketitle

\begin{abstract}
Hyper-heuristics have become a popular approach for solving dynamic flexible job shop scheduling (DFJSS) problems. They use gradient-free optimization techniques like Genetic Programming (GP) to evolve non-differentiable heuristics. However, conventional GP methods tend to converge slowly because they rely solely on evolutionary search to find good heuristics. Existing multitask GP methods can solve multiple tasks simultaneously and speed up the search by transferring knowledge across similar tasks. But they mostly exchange heuristic building blocks without truly generating heuristics conditioned on task information. 
In this paper, we aim to accelerate convergence and enable task-specific heuristic generation by incorporating a task-conditioned Transformer model. The Transformer works in two ways. First, it learns the distribution of elite heuristics, biasing the search toward promising regions of the heuristic space. Second, through conditional generation, it produces heuristics tailored to specific tasks, allowing the model to handle multiple scheduling tasks at once and improving overall optimization efficiency. 
Based on these ideas, we propose TransGP, a Task-Conditioned Transformer-Guided GP framework. This evolutionary paradigm integrates generative modeling with GP, enabling efficient multitask heuristic learning and knowledge transfer. We evaluate TransGP on a range of DFJSS scenarios. Experimental results show that TransGP consistently outperforms multitask GP baselines, widely used handcrafted heuristics, and the pure Transformer model, achieving faster convergence, superior solution quality, and enhanced robustness.
\end{abstract}


%
\IEEEpeerreviewmaketitle

\section{Introduction}
\label{sec:introduction}
\IEEEPARstart{S}{cheduling} is a fundamental problem across domains like intelligent manufacturing and supply chains \cite{ding2019two}. 
In this paper, we focus specifically on the dynamic flexible job shop scheduling (DFJSS) problem, where jobs arrive over time and each operation can be processed on one of several alternative machines with machine-dependent processing times under continuous shop-floor disturbances \cite{xu2023genetic}. While many classical approaches have been developed over the past decade \cite{durasevic2023heuristic, corsini2024self}, their high computational cost limits scalability due to the NP-hard nature of the problem. 
Recently, the rise of artificial intelligence has opened new avenues through learning-based optimization \cite{wang2021bi}. These approaches fall into two main categories \cite{xu2025learn}: neural combinatorial optimization via reinforcement learning \cite{yi2025improved}, and symbolic regression-based hyper-heuristics \cite{xu2025quality}. In line with industrial requirements for human-interpretable decision support in DFJSS, this paper focuses on the latter for its simple framework and strong interpretability \cite{mei2022explainable}. 

Symbolic regression-based hyper-heuristics aim to automatically learn explicit and interpretable rules for job sequencing and routing in DFJSS problems. Genetic Programming (GP) \cite{zhong2018multifactorial} has proven to be an effective approach for evolving such symbolic scheduling heuristics \cite{zhou2020automatic}. However, as a gradient-free method, GP often suffers from low convergence efficiency, typically requiring prolonged training to discover high-quality heuristics. To improve the convergence of GP, we consider incorporating generative models into the evolutionary process, motivated by two key considerations.

\begin{figure*}[t]
\centering
\includegraphics[width=0.99\linewidth]{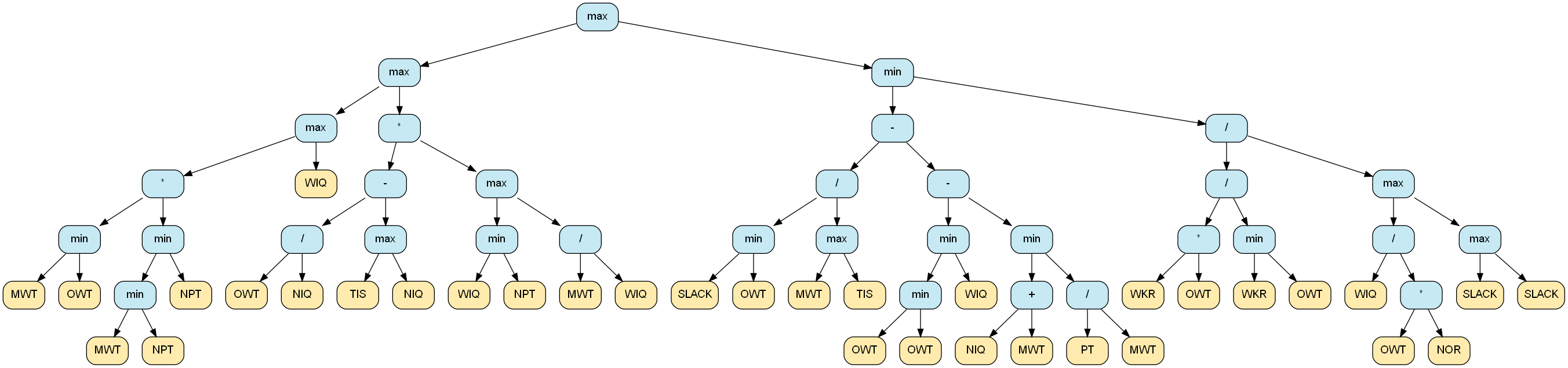}
\caption{An example of a symbolic heuristic in DFJSS, which contains complex and irregular structures.}
\label{fig:example_heuristic}
\end{figure*}

First, the inefficiency of GP largely arises from the limitations of traditional evolutionary operators (e.g., crossover and mutation), which depend heavily on random variation. Although such operators enable broad exploration, they are inherently simplistic and often fail to capture or preserve the complex, irregular structures that characterize effective subtrees in symbolic heuristic spaces. For example, high-performing DFJSS heuristics frequently contain deeply nested conditional subtrees, nonlinear interactions among multiple job- and machine-level features, and hybrid constructs that combine workload, slack, queue length, processing time, and other attributes in nonstandard patterns (see Fig.~\ref{fig:example_heuristic}). These structural motifs do not follow simple algebraic templates and rely on coordinated dependencies across multiple subtrees, making them highly vulnerable to disruption by random crossover or mutation. This leads to the fact that a large number of offspring heuristics generated by evolutionary operators exhibit low fitness, resulting in wasted evaluations and slow convergence. To overcome this limitation, a promising direction is to integrate a generative model \cite{guidotti2024generative, zhang2025llm} into the GP process, allowing the model to learn the underlying distribution of elite heuristics and guide the search toward higher-quality regions of the heuristic space. By learning these structural motifs directly from elite heuristics, the generative model can reproduce nontrivial subtrees, preserve latent dependencies, and generate pattern-conforming structures that standard GP operators would rarely discover or maintain.

Second, real-world optimization problems rarely exist in isolation. In DFJSS, families of related tasks commonly arise from varying utilization levels, arrival patterns, disturbance intensities, or due-date tightness \cite{luo2026knowledge}. These tasks share decision structures (machine assignment and sequencing) but differ in stochastic characteristics, creating a natural multitask optimization setting. Conditional generative models can exploit these variations by guiding the optimization of multiple related scheduling tasks in a problem-aware manner, thereby further improving efficiency across task instances.

Taken together, these considerations motivate the integration of generative modeling into GP to enhance its ability to produce high-quality symbolic heuristics more effectively and adaptively. Building on these insights, we propose Task-Conditioned Transformer-guided GP (TransGP), a novel framework that integrates the generative modeling capabilities of Transformers \cite{vaswani2017attention} with the adaptive, population-based search of GP. The Transformer component enables the learning of the underlying distribution of high-quality heuristics, guiding the generation process toward promising regions of the search space. Furthermore, by incorporating task-specific information, the model supports conditional generation, allowing GP to generalize across diverse scheduling tasks parametrized by different environment settings. In this work, these tasks correspond to multiple DFJSS environments that share the same decision structure but differ in machine workloads and dynamic shop-floor conditions. Our core contributions are summarized as follows:
\begin{itemize}
    \item We propose a new mechanism to encode symbolic GP individuals (tree-structured heuristics) into Transformer-compatible vector sequences and decode generated sequences back into valid expression trees. This enables the Transformer to model complex structural motifs and long-range dependencies that conventional GP operators cannot capture.
    \item The Transformer is trained to model elite heuristic distributions, enabling it to serve as a semantic-aware mutation operator that generates pattern-conforming, structurally meaningful subtrees instead of random modifications. This substantially enhances the efficiency and quality of symbolic heuristic evolution.
    \item We incorporate task-specific embeddings into the Transformer, allowing the generated heuristics to adapt to different scheduling tasks based on the embedded task information. This conditional generation enables TransGP to transfer structural knowledge across related tasks while still producing specialized heuristics for each scenario.
    \item Extensive experiments across diverse DFJSS scenarios demonstrate that TransGP consistently outperforms multitask GP baselines, widely used handcrafted heuristics, and pure Transformer, achieving faster convergence, superior heuristic quality, and better interpretability within the DFJSS problem class.
\end{itemize}

The remainder of this paper is organized as follows. Section \ref{relatedWork} reviews related work. Section \ref{method} details the proposed TransGP framework. The experimental design is described in Section \ref{sec:design}, and the experimental results are presented in Section \ref{sec:results}. Section \ref{analysis} provides an in-depth analysis of the evolved heuristics, and Section \ref{conclusion} concludes the paper.

\section{Related Work}
\label{relatedWork}

\subsection{Multitask Dynamic Flexible Job Shop Scheduling}
The Dynamic Flexible Job Shop Scheduling (DFJSS) problem extends the classical job shop scheduling \cite{zhang2019review} framework by incorporating \textit{machine flexibility} and \textit{dynamic environmental uncertainties}, such as stochastic job arrivals \cite{lei2023large}.  
Let $\mathcal{J} = \{J_1, J_2, \dots, J_n\}$ denote a set of jobs, where each job $J_i$ is characterized by an arrival time $r_i$, a weight $\rho_i$, a due date $d_i$, and a sequence of operations $\{O_{i1}, O_{i2}, \dots, O_{im_i}\}$.  
Each operation $O_{ik}$ can be processed by a subset of machines $\mathcal{M}_{ik} \subseteq \mathcal{M} = \{M_1, M_2, \dots, M_h\}$, with potentially different processing times $p_{ik}^{(j)}$ on machine $M_j \in \mathcal{M}_{ik}$.  
When machines are geographically distributed, a transportation time $\tau_{k_1,k_2}$ is incurred when transferring a job between machines $M_{k_1}$ and $M_{k_2}$. The scheduling process must satisfy the following constraints:
\begin{itemize}
	\item Operation precedence: Operations within a job must follow a predefined sequence. Operation $O_{ij}$ can start only after $O_{i(j-1)}$ is completed, if $j > 1$.
	\item Non-preemption: Once an operation starts on a machine, it must run to completion before the machine can be reassigned.
	\item Machine capacity: Each machine $M_k$ can process at most one operation at any given time.
	\item Machine assignment: Each operation $O_{ij}$ must be assigned to exactly one machine from its eligible machine set $\mathcal{M}_{ij}$.
\end{itemize}

Real manufacturing systems rarely operate under a single, fixed performance criterion. Different production environments, such as high-urgency orders or small number of machines, require \textit{distinct task-specific scheduling heuristic} \cite{chen2025optimizing}. This motivates a \textit{multi-task} optimization setting, where a single learning framework is trained to handle multiple related scheduling tasks, each defined by different shop floor settings, while leveraging shared knowledge across tasks.

In this study, each task corresponds to a specific shop floor setting (i.e., different utilization levels and number of machines), and a task-conditioned scheduling heuristic is trained to optimize all tasks jointly.  
We focus on three widely adopted objectives \cite{xu2023genetic}:
\begin{itemize}
	\item \(\displaystyle F_{\text{max}} = \max_{i=1}^n (c_i - r_i)\): maximum flowtime, measuring the worst-case job turnaround.
	\item \(\displaystyle F_{\text{mean}} = \frac{1}{n}\sum_{i=1}^n (c_i - r_i)\): mean flowtime, reflecting overall system responsiveness.
	\item \(\displaystyle T_{\text{mean}} = \frac{1}{n}\sum_{i=1}^n \max\bigl(0, c_i - d_i\bigr)\): mean tardiness, quantifying average delay relative to due dates.
\end{itemize}
Here, $c_i$ denotes the completion time of job $J_i$, $r_i$ its release time, and $d_i$ its due date.

Studying DFJSS in a multi-task framework is crucial for several reasons:
\begin{enumerate}
	\item Knowledge transfer across tasks \cite{zhang2022task}: Related scheduling tasks often share structural patterns such as machine conflicts or bottleneck dynamics. Multi-task learning can exploit these commonalities to improve generalization and reduce training cost compared with optimizing each task independently.
	\item Robust decision making \cite{chen2024generate}: A model trained across diverse tasks can produce more flexible scheduling policies that adapt to changing business priorities or production conditions.
	\item Practical deployment \cite{zhang2021surrogate}: In real manufacturing systems, shop-floor environments may change over time (e.g., transitioning from normal operation to high-load rush periods). Multi-task optimization offers a unified policy space that accommodates such dynamic changes without the need for retraining from scratch.
\end{enumerate}

In the experiments, multiple DFJSS tasks are defined to evaluate the proposed method’s ability to learn generalizable scheduling heuristics that remain effective across diverse manufacturing conditions.

\subsection{Symbolic Heuristics for Scheduling}
Symbolic rule-based heuristics such as dispatching rules \cite{rajendran1999comparative} and composite priority functions \cite{teymourifar2022comparison} have a long history in job shop and flow shop scheduling due to their low computational overhead and domain interpretability \cite{chen2013flexible}. However, designing effective rules typically requires expert knowledge and extensive manual tuning, which limits their adaptability to new problem instances or environments \cite{xu2025learn}. To address this limitation, researchers have employed GP to automate the discovery of scheduling heuristics \cite{braune2022genetic, zeitrag2024cooperative, shady2020automatic}. GP evolves symbolic expressions in the form of expression trees, enabling the synthesis of interpretable rules directly from task data \cite{xu2025quality, koza1999genetic}. Early work demonstrated the viability of GP for evolving scheduling rules in job shop and flow shop environments \cite{shady2022novel, nguyen2012computational, guo2024improved}. Nevertheless, despite these successes, conventional GP approaches still suffer from low convergence efficiency and typically require numerous evaluations to obtain high-quality heuristics.

\subsection{Genetic Programming for Multitask Scheduling}
Multitask optimization in GP aims to evolve a population of heuristics that generalize across a family of related scheduling tasks \cite{zhang2022task, gupta2015multifactorial}. This paradigm leverages structural similarities among tasks to accelerate convergence and enhance generalization performance \cite{zhang2022task}. Prior approaches have introduced techniques such as surrogate modeling \cite{zhang2021surrogate}, co-evolutionary strategies \cite{chen2025optimizing}, and fitness sharing to promote diversity and enable inter-task knowledge transfer. Some also estimate task-relatedness to guide transfer and adaptation \cite{zhang2022task}. These approaches demonstrate that leveraging knowledge from related tasks can indeed improve convergence efficiency, enabling GP to discover high-quality scheduling heuristics with fewer evaluations. However, most existing multitask GP methods still rely primarily on traditional evolutionary operators such as crossover and mutation. Because these operators depend heavily on randomness, they often fail to capture the distribution of high-quality solutions in the heuristic space, leading to a high proportion of low-quality offspring. Furthermore, practical scheduling scenarios are often characterized by explicit environmental parameters, such as demand fluctuations or shop floor congestion levels, that could provide valuable contextual information for guiding evolutionary search. However, existing multitask GP approaches frequently overlook such information. Together, these limitations highlight considerable scope for improving the convergence efficiency of multitask GP.


\subsection{Generative Models for Symbolic Expression Learning}
In scheduling domains, deep learning has been widely adopted for scheduling policy learning \cite{zhang2020learning, zhang2024deep, kotary2022fast}, but most models rely on opaque, black-box representations that limit interpretability \cite{xu2025learn}. Recent ongoing work has sought to bridge this gap by leveraging generative models to evolve symbolic rules \cite{chen2025genetic, romera2024mathematical, liu2024evolution, dat2025hsevo, novikov2025alphaevolve, ye2024reevo}. In particular, Transformer \cite{vaswani2017attention} architectures have achieved strong performance in symbolic domains such as program synthesis \cite{tao2023program}, theorem proving \cite{gontier2020measuring}, and molecular generation \cite{bagal2021molgpt}. These models excel at capturing structural regularities in symbolic sequences and have been used to encode domain knowledge within neural-guided search frameworks \cite{han2021transformer}. However, most existing methods focus on single-task settings and do not address the unique challenges of generating task-aware symbolic heuristics for multitask optimization. 

Motivated these gaps, our work proposes a task-conditioned symbolic generation framework based on Transformer models. By embedding task-specific information into the generation process, the model produces semantically meaningful symbolic expressions tailored to the unique characteristics of each task. Unlike traditional multitask GP approaches that rely on undirected transfer, our framework enables guided multitask optimization, where the Transformer acts as a learned mutation operator, transferring structural knowledge across tasks while maintaining task-specific adaptability through joint training on diverse task archives



\section{Method}
\label{method}
\begin{figure*}[t]
    \centering
    \includegraphics[width=0.99\linewidth]{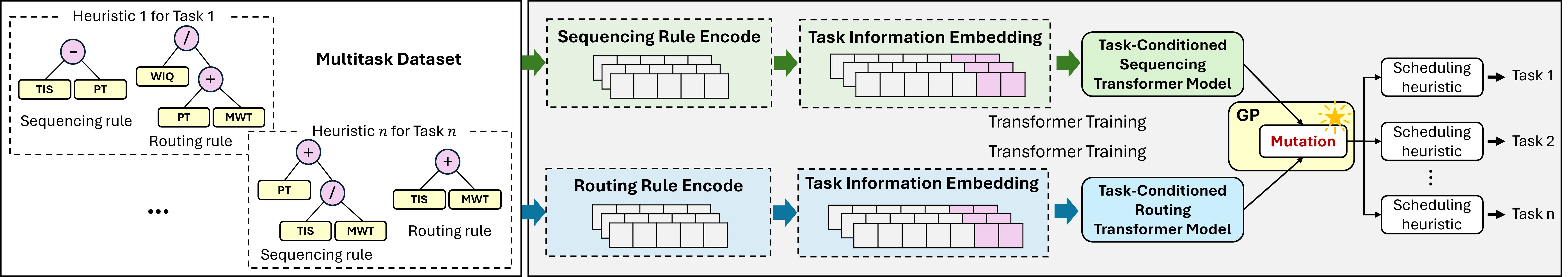}
    \caption{Overall framework of TransGP.}
    \label{fig:overallframework}
\end{figure*}

\subsection{Overall Framework}

The core of TransGP lies in its generative approach to evolving symbolic heuristics. We formulate the generation of routing and sequencing rules as a sequence modeling problem by linearizing their abstract syntax trees into token sequences. This representation naturally aligns with the Transformer architecture, which excels at capturing complex dependencies in structured sequential data. Leveraging this strength, we train Transformer models to learn the underlying distribution of high-quality heuristics. This learning-based approach provides two key advantages: (1) it models the complex and irregular distribution of elite heuristics more effectively than stochastic operators like crossover and mutation, and (2) it enables multitask optimization by incorporating task-conditioned embeddings, allowing for the conditional generation of heuristics tailored to specific tasks.

Fig.~\ref{fig:overallframework} illustrates the TransGP framework. The process starts with a dataset of elite heuristics collected from various DFJSS tasks. In principle, this dataset can come from any DFJSS scenarios, regardless of whether the heuristics are handcrafted by experts or discovered through GP methods, as long as they share the same heuristic space (i.e., the same terminal and function sets). This aligns with real-world practice: there often already exist multiple good heuristics trained for similar DFJSS scenarios. Consider a shop floor where conditions shift, say due to seasonal changes. The new task may call for a more specific heuristic, but we would rather not train from scratch. Instead, we want to leverage prior knowledge and speed up the process.

In this paper, we collect elite heuristics via GP across multiple scheduling tasks as a demonstration. The details of this dataset construction strategy are provided in Section~\ref{sec:parameter_setting}. From this dataset, we train two task-conditioned Transformer models: one for sequencing and one for routing. Each heuristic is converted into vector representations using dedicated modules, while task embeddings serve as conditioning inputs to capture task-specific patterns. These embeddings allow the Transformers to generate heuristics adapted to the unique characteristics of each task. Once trained, the Transformer models are integrated into the GP pipeline as mutation operators. During evolution, they generate new, task-aware heuristic candidates, steering the search toward compact and high-performing heuristics. This guided mutation improves both the efficiency and the effectiveness of the evolutionary process. 

Overall, the proposed method consists of two offline stages. The first stage trains the two Transformer models. The second stage trains the Transformer-guided GP. The final heuristics obtained can then be applied online to make real-time decisions whenever a decision point occurs. The following sections describe the task-conditioned Transformer for symbolic heuristic generation and the Transformer-guided GP framework, which together form the complete TransGP approach.

\begin{figure}[t]
    \centering
    \includegraphics[width=0.90\linewidth]{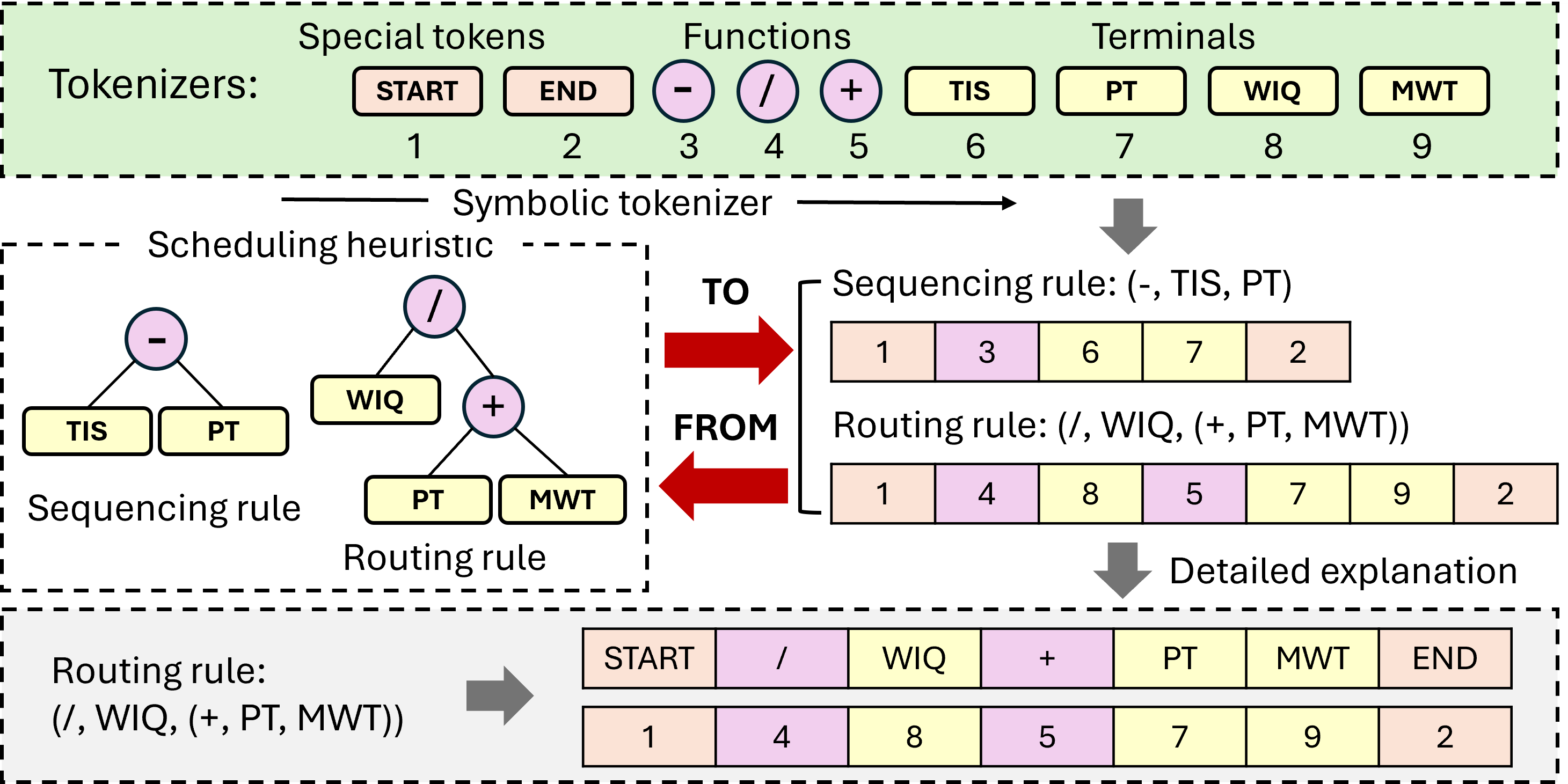}
    \caption{Vectorization of symbolic heuristics.}
    \label{fig:encode-decode}
\end{figure}

\subsection{Task-Conditioned Transformer Model for Symbolic Heuristic Generation}

\subsubsection{Vectorization of Symbolic Heuristics}
Fig.~\ref{fig:encode-decode} illustrates the process of converting symbolic heuristics to and from a vector representation. Each scheduling heuristic has two trees: one representing the sequencing rule and the other the routing rule. Internal nodes correspond to \textit{functions} $A = \{a_1, a_2, \ldots, a_m\}$, while leaf nodes correspond to \textit{terminals} $B = \{b_1, b_2, \ldots, b_n\}$. In our setup, all functions are binary. For example, the sequencing rule \texttt{-(TIS, PT)} is represented as a tree with \texttt{-} as the root node and \texttt{TIS} and \texttt{PT} as its children. To enable training with a Transformer, symbolic expression trees are linearized using a prefix (pre-order) traversal. Special tokens \texttt{START} and \texttt{END} are added and appended, respectively, to denote the sequence boundaries. Formally, given a symbolic expression tree $S$, it is converted into a node sequence $\mathbf{n} = [n_1, n_2, \ldots, n_L]$, where $n_1 = \texttt{START}$, $n_L = \texttt{END}$, and each intermediate node $n_i \in A \cup B$ for $1 < i < L$. Each node $n_i$ is assigned a unique index based on a predefined mapping, resulting in a discrete input sequence:
\begin{equation}
    \mathbf{x} = [x_1, x_2, \ldots, x_L] = [\text{ID}(n_1), \text{ID}(n_2), \ldots, \text{ID}(n_L)]
\end{equation}
This sequence $\mathbf{x}$ serves as the input for training the Transformer-based model.

\subsubsection{Task-Conditioned Transformer Training}
To support task-conditioned generation of scheduling heuristics, we implement a Transformer decoder model that autoregressively generates prefix-form token sequences. Given an input sequence of tokens $\mathbf{x} = [x_1, x_2, \dots, x_L] \in \mathbb{N}^L$, the model first computes token embeddings $\mathbf{E}_x = \text{Embed}(\mathbf{x}) \in \mathbb{R}^{L \times d}$ and adds learnable positional encodings $\mathbf{P} \in \mathbb{R}^{L \times d}$, where $d$ is the model dimension. For given task-specific conditional parameters $\mathbf{e}_{\text{task}} \in \mathbb{R}^{d_{\text{task}}}$, where $d_{\text{task}}$ is the dimension of the task-specific conditional parameter, we apply a linear projection $\mathbf{C} = \text{linear}(\mathbf{e}_{\text{task}}) \in \mathbb{R}^d$, and then expand it across the sequence dimension to obtain a conditioning tensor $\mathbf{C}_{\text{seq}} \in \mathbb{R}^{L \times d}$. The total input of the decoder is then:
\begin{equation}
    \mathbf{H}_0 = \mathbf{E}_x + \mathbf{P} + \mathbf{C}_{\text{seq}}.
\end{equation}
The decoder comprises a stack of $Q$ Transformer decoder layers. Each layer consists of masked self-attention and position-wise feedforward submodules, enabling autoregressive sequence generation. The hidden representation at the $q$-th layer is computed as:
\begin{equation}
    \mathbf{H}_q = \mathrm{DecoderLayer}_q(\mathbf{H}_{q-1}, \mathrm{mask}), \; q = 1, \dots, Q,
\end{equation}
where the memory input is set to zero, and a causal mask enforces the autoregressive constraint. Each Transformer decoder layer comprises an attention module followed by a multilayer perceptron module. Within each attention module, the input is projected into query $\textbf{Q}$, key $\textbf{K}$, and value $\textbf{V}$ matrices via learned projections $\mathbf{W}_Q, \mathbf{W}_K, \mathbf{W}_V$, respectively. The scaled dot-product attention is then computed as:
\begin{equation}
    \mathrm{Attention}(\mathbf{Q}, \mathbf{K}, \mathbf{V}) = \mathrm{softmax}\left( \frac{\mathbf{Q} \mathbf{K}^\top}{\sqrt{d_k}} \right) \mathbf{V}.
\end{equation}
To more precisely capture the features of elite heuristics, our attention module adopts a multi-head self-attention mechanism, enabling the model to attend to different representation subspaces for multiple tasks jointly:
\begin{equation}
    \tilde{\mathbf{M}} = \mathrm{Concat}(\mathrm{head}_1, \dots, \mathrm{head}_H) \mathbf{W}^{(O)},
\end{equation}
where $\mathrm{head}_i$ denotes the output of the $i$-th attention head, and $\mathbf{W}^{(O)}$ is the output projection matrix.
Finally, the output logits are computed by applying a linear transformation to the final decoder layer outputs, producing a distribution over all possible node types:
\begin{equation}
    \hat{\mathbf{Y}} = \mathbf{H}_Q \mathbf{W}_{\mathrm{out}}^\top + \mathbf{b}_{\mathrm{out}}, \quad \hat{\mathbf{Y}} \in \mathbb{R}^{L \times R},
\end{equation}
where $L$ is the sequence length and $R$ is the rule size, including all functions, terminals, and special tokens.


To train the model, we minimize the standard cross-entropy loss over the token sequence, based on a dataset of elite heuristics denoted by $\mathcal{D} = \{ (\mathbf{x}^{(u)},\textbf{e}_{\text{task}}^{(u)}) \}_{u=1}^{U}$. Here, $\textbf{x}^{(u)}$ represents the vectorized heuristic, and $\textbf{e}_{\text{task}}^{(u)}$ denotes the corresponding task-specific conditional parameters for the $u$-th sample in the dataset, with $U$ indicating the total number of samples. Let $p(\cdot)$ denote the probabilistic distribution defined by the Transformer decoder model, and define $\textbf{E}_{x}^{(u)} = \text{Embed}(\textbf{x}^{(u)})$. The loss function is given by:
\begin{equation}
\mathcal{L} = - \frac{1}{U \cdot L} \sum_{u=1}^{U-1} \sum_{i=1}^{L-1} \log p({x}^{(u)}_{i+1} \mid \textbf{E}^{(u)}_{x,1:i}, \textbf{e}^{(u)}_{\text{task}}),
\end{equation}
where $L$ is the sequence length, $x^{(u)}_{i+1}$ is the $(i+1)$th component of $\textbf{x}^{(u)}$, and $\textbf{E}_{x,1:i}$ refers to the submatrix consisting of the first $i$ columns of $\textbf{E}_{x}$. This objective encourages the model to generate valid and task-relevant sequencing/routing rules by maximizing the likelihood of the correct next {token} at each position. The trained task-conditioned Transformer models for sequencing and routing are then integrated into the GP mutation pipeline to enable guided mutation.


\subsection{Task-Conditioned Transformer-Guided GP}
\label{sec:mutation}
\begin{figure}[t]
    \centering
    \includegraphics[width=0.99\linewidth]{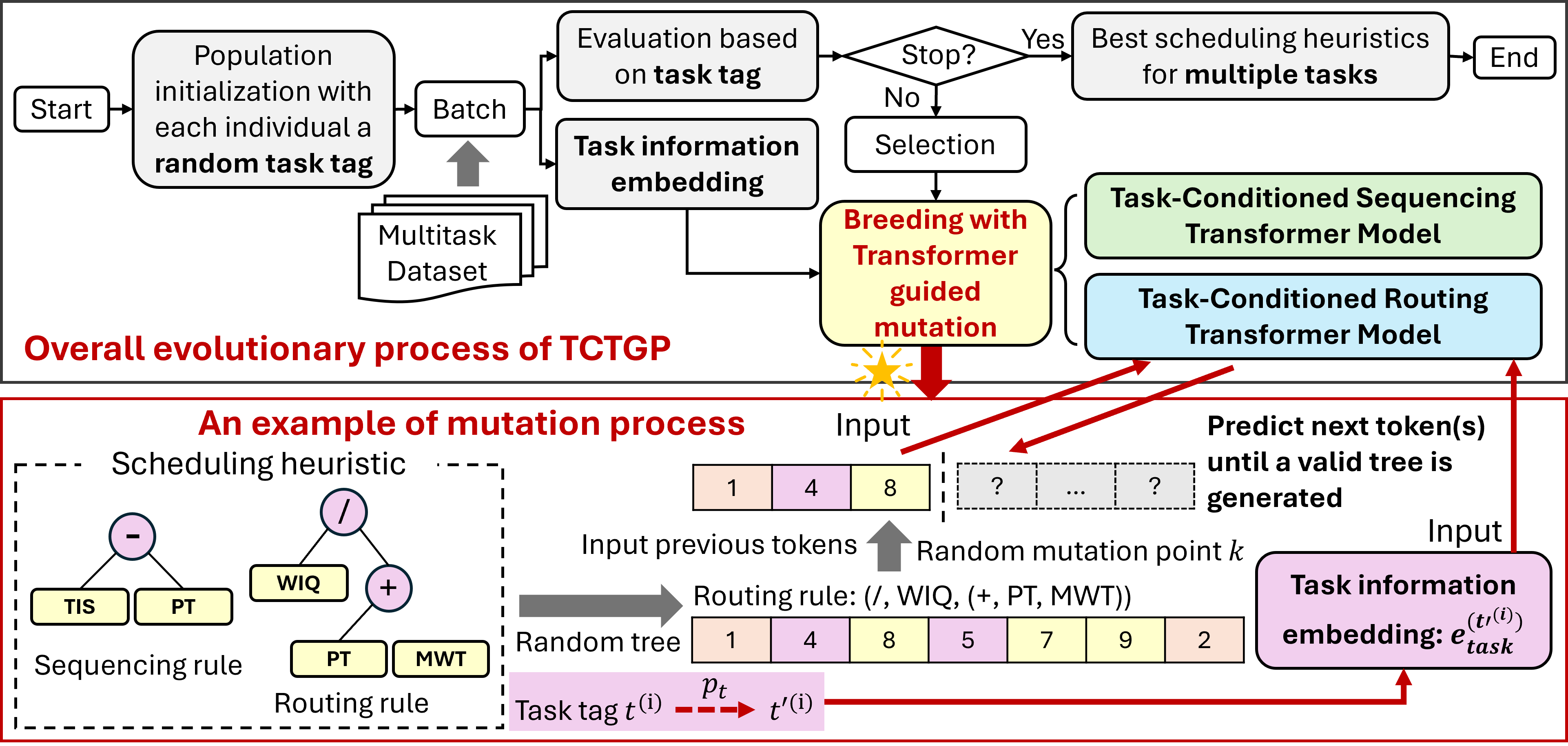}
    \caption{Overview of the evolutionary framework in TransGP and an example of Transformer-guided mutation process.}
    \label{fig:transGP}
\end{figure}

The overall evolution process of TransGP, incorporating the learned Transformer, illustrated in Fig.~\ref{fig:transGP}, begins by initializing a population of heuristic individuals, each assigned a random target task. Individuals are evaluated solely on their assigned tasks using batches of multitask instances, and fitness scores are computed. Task embeddings are generated in parallel. Top individuals are selected as parents and undergo Transformer-guided mutation, where task embeddings inform the generation of offspring. These offspring replace the current population, and the process repeats until a stopping criterion is met, yielding optimized heuristics across tasks. At iteration $g$, the GP population is:
\begin{equation}
    \mathcal{P}^{(g)} = \{ (h^{(i)}, t^{(i)}) \}_{i=1}^{N},
\end{equation}
where ${h}^{(i)}$ is a symbolic scheduling heuristic and $t^{(i)}$ denotes the corresponding task. Each individual $({h}^{(i)}, t^{(i)})$ is evaluated solely on the task $t^{(i)}$. The fitness of an individual is computed as:
\begin{equation}
    f(h^{(i)}, t^{(i)}) = \text{Obj}(h^{(i)}, t^{(i)}),
\end{equation}
where $\text{Obj}(h^{(i)}, t^{(i)})$ denotes the task-specific performance metric (e.g., mean flowtime or mean tardiness) on task $t^{(i)}$. This formulation enables TransGP to evolve specialized heuristics across diverse tasks in parallel, providing a scalable foundation for multitask evolutionary framework.


\subsubsection{Task-Conditioned Transformer-Guided Mutation}
As mentioned, the core innovation of our approach lies in this task-conditioned Transformer-guided mutation mechanism, which leverages pretrained task-conditioned Transformer models to perturb parent individuals in a task-aware manner. As illustrated in Fig.~\ref{fig:transGP}, mutation proceeds as follows. Given a parent individual $(h^{(i)}, t^{(i)})$, the mutation process may optionally shift the task focus to a different task $t'^{(i)} \ne t^{(i)}$, sampled uniformly from the set of alternative tasks with probability $p_t$. If a task switch occurs, the corresponding task embedding $\textbf{e}^{(t'^{(i)})}_{\text{task}}$ is used instead. This mechanism enables cross-task knowledge transfer during the evolution process. Next, mutation targets either the sequencing or routing rule of the individual’s scheduling heuristic. A random mutation point $k$ is selected for modification; let $x_{k:}$ denote the prefix-ordered tokens from the mutation point to the end of this tree. To ensure syntactic validity, a stack-based mechanism is used, maintaining well-formed prefix expressions throughout the process. The symbolic heuristic is then generated step by step as follows:
\begin{itemize}
    \item Compute the raw logits: $logits_{x_{k:}}$.
    \item Maintain a \texttt{stack} to track the number of children expected for open function nodes.
    \item Based on the current stack, determine a set of valid tokens $\mathcal{V}_{\text{valid}}$. For example, if the current node requires children, both functions and terminals are allowed; otherwise, only \texttt{END} is valid.
    \item Mask all invalid tokens by setting their logits to $-\infty$:
    \begin{equation}
        logits_{\text{masked}}[j] = \begin{cases} 
    logits_{\text{raw}}[j] & \text{if } \text{ID}^{-1}(j) \in \mathcal{V}_{\text{valid}} \\
    -\infty & \text{otherwise} 
    \end{cases}
    \end{equation}
    \item Sample the next token $x_{k+1}$ from the softmax distribution:
    \begin{equation}
    p_{k+1} = \text{softmax}\left(\frac{logits_{\text{masked}}}{\gamma}\right),
    \end{equation}
    where $\gamma$ is a temperature parameter that controls the exploration-exploitation trade-off. Higher values of $\gamma > 1.0$ encourage more diverse mutations, while lower values $\gamma < 1.0$ result in greedier, more deterministic choices.
    \item Update the stack: if a function is chosen, push 2 onto the stack and decrement the current top; if a terminal is chosen, decrement and pop if zero.
\end{itemize}

This procedure ensures that the generated subtrees are both syntactically valid and semantically aligned with the target task. The resulting mutated individual is then evaluated on its assigned task, and passed into the next generation.

\subsection{Computational Cost Discussion}
\label{sec:cost}
This section analyzes the computational overhead introduced by TransGP and clarifies its impact on the overall evolutionary process. The cost arises from two sources: offline training of the Transformer and its online use during genetic programming evolution.

The Transformer models are trained offline prior to evolution, using a limited set of elite heuristics collected from multiple scheduling tasks. These models are lightweight and trained only once per scenario, after which they are reused throughout the evolutionary search. Consequently, the training cost does not scale with the number of generations and does not accumulate during GP evolution. During evolution, the Transformer is invoked solely for guided mutation, where it autoregressively generates a symbolic subtree to replace part of an existing heuristic. Given that all functions are binary and the tree depth is bounded, the length of the generated sequence is strictly limited. Let $L$ denote the number of generated tokens, $d$ the number of decoder layers, and $Q$ the hidden dimension; the computational complexity of a single guided mutation is $O(L d Q^2)$, with $L$ bounded by the maximum tree depth. In practice, the generated subtrees are not huge, keeping inference costs modest. At the system level, the overall runtime is dominated by simulation-based fitness evaluation of DFJSS heuristics, which is substantially more computationally demanding than GP variation operators. As a result, the additional overhead from Transformer-guided mutation constitutes only a minor portion of the total runtime and does not change the primary computational bottleneck.

Overall, TransGP introduces limited and well-controlled computational overhead. The offline training cost is amortized across the entire run, and the online inference cost is negligible relative to fitness evaluation, enabling improved search performance without materially increasing end-to-end computational cost.


\section{Experiment Design}
\label{sec:design}
\subsection{DFJSS Simulator}
To evaluate the proposed approach, we generate DFJSS scenarios using a discrete-event simulation model that emulates a DFJSS environment \cite{xu2025learn}. The simulated system includes 10 heterogeneous machines processing 124 jobs, with machine speeds uniformly sampled from $[10, 15]$. Transportation times between machines and to entry/exit points follow a discrete uniform distribution over $[7, 100]$. Jobs arrive via a Poisson process to reflect real-time scheduling. Each job comprises 2–10 operations, with operation workloads drawn uniformly from $[100, 1000]$. Due dates are set to 1.5 times the job’s total processing time. To promote generalization, we use a single simulation replication per scenario but introduce variability across generations by randomizing the seed \cite{hildebrandt2010towards}. We evaluate performance under three machine configurations (6, 8, and 10), each associated with system utilization levels of 0.75, 0.85, and 0.95, representing increasing scheduling difficulty. Each experimental scenario consists of three tasks defined by a specific scheduling objective and system configuration. Tasks are named using the format \texttt{<obj-utilLevel-machNum>}, where \texttt{obj} is the scheduling goal (e.g., \texttt{Fmax}, \texttt{Fmean}, \texttt{Tmean}). Table~\ref{tab:scenarios} summarizes all scenarios and tasks.

\begin{table}[t]
\footnotesize
\centering
\caption{Summary of scenarios and corresponding tasks}
\label{tab:scenarios}
\setlength{\tabcolsep}{1mm}
\begin{tabular}{ccccc}
\hline
\textbf{S*} & \textbf{Task} & \textbf{Obj} & \textbf{UtilLevel} & \textbf{MachNum} \\ \hline
\multirow{3}{*}{\textbf{1}} 
& Task 1: <Fmax-0.75-6>   & Fmax  & 0.75 & 6 \\
& Task 2: <Fmax-0.85-8>   & Fmax  & 0.85 & 8 \\
& Task 3: <Fmax-0.95-10>  & Fmax  & 0.95 & 10 \\ 
\hline
\multirow{3}{*}{\textbf{2}} 
& Task 1: <Fmean-0.75-6>  & Fmean & 0.75 & 6 \\
& Task 2: <Fmean-0.85-8>  & Fmean & 0.85 & 8 \\
& Task 3: <Fmean-0.95-10> & Fmean & 0.95 & 10 \\ 
\hline
\multirow{3}{*}{\textbf{3}} 
& Task 1: <Tmean-0.75-6>  & Tmean & 0.75 & 6 \\
& Task 2: <Tmean-0.85-8>  & Tmean & 0.85 & 8 \\
& Task 3: <Tmean-0.95-10> & Tmean & 0.95 & 10 \\ 
\hline
\end{tabular}
\end{table}

\subsection{Parameter Setting}
\label{sec:parameter_setting}
The terminal and function sets used to construct heuristic individuals are summarized in Table~\ref{notation}. Terminals describe dynamic system features at the machine, operation, and job levels \cite{xu2025learn}. Consistent with prior GP-based DFJSS studies \cite{xu2025learn, djurasevic2016adaptive, zhang2023survey}, we adopt a standard, empirically validated terminal set that has been shown to provide sufficient and robust information for effective sequencing and routing decisions. These terminals capture core operational factors, such as machine congestion, job urgency, and remaining processing requirements, and serve as a shared feature basis across tasks, which is critical for multi-task learning and cross-task knowledge transfer in TransGP. The function set consists of basic binary arithmetic operators, with protected division returning 1 when the divisor is 0. 

All methods use a population size of 600 and evolve heuristics over 50 generations, with ramped half-and-half initialization. Parent selection is performed using tournament selection (size = 5), and elitism preserves the top 4 individuals per generation. Crossover is disabled so that mutation is the sole variation operator, allowing us to isolate the effect of Transformer-guided mutation. 
The Transformer is trained on an elite-heuristic dataset collected across multiple generations and tasks, following a multi-task, multi-generation data construction strategy. The resulting dataset comprises 3,600 elite heuristics (population size 600, top 20 elites per generation, collected over the last 20 generations, 3 tasks, and 3 independent runs), and is further filtered to remove duplicate heuristics. The Transformer is trained on a CPU. It has 256 hidden dimensions, 4 attention heads, 4 layers, and a dropout rate of 0.1. During evolution, Transformer-guided mutation is applied stochastically using temperature-controlled sampling to preserve exploration, ensuring that learned structural priors guide, rather than dominate, the GP search process.

\begin{table}[t]
\centering
\footnotesize
\caption{The terminal and function set.}
\label{notation}
\begin{tabular}{cl}
\hline
\textbf{Notation} & \multicolumn{1}{c}{\textbf{Description}}                    \\ 
\hline
\texttt{NIQ}               & Number of operations in the queue                       \\
\texttt{WIQ}               & Work in the queue                                       \\
\texttt{MWT}               & Waiting time of the machine                              \\
\texttt{PT}                & Processing time of the operation                         \\
\texttt{NPT}               & Median processing time for the next operation           \\
\texttt{OWT}               & Waiting time of the operation                           \\
\texttt{WKR}               & Work remaining                                          \\
\texttt{NOR}               & Number of operations remaining                          \\
\texttt{SLACK}             & Slack                                                   \\
\texttt{TIS}               & Time in system = current time - release time            \\
\hline
Function          & $+$, $-$, $*$, \text{protected} $/$, $\max$, $\min$              \\
\hline
\end{tabular}
\end{table}

\subsection{Comparison Design}
To assess the effectiveness of TransGP, we compare it against two GP baselines: (i) GP, which employs standard mutation without task conditioning, and (ii) TGP, an extension of \cite{chen2025optimizing} that introduces probabilistic task-tag mutation to enable inter-task transfer while retaining conventional GP mutation operators. This comparison isolates the contribution of Transformer-guided, task-conditioned mutation to heuristic discovery. To further evaluate practical effectiveness, TransGP is compared with widely used handcrafted heuristics as well as a Transformer-only baseline that generates heuristics without evolutionary search. These comparisons allow us to disentangle the benefits of the proposed framework from those of GP-based evolutionary search and to assess the necessity of evolutionary refinement when using learned generative models. To ensure a fair comparison, each method is executed for 30 independent runs. The best scheduling heuristic learned from each run, along with selected handcrafted heuristics, is then evaluated on a separate test set. The resulting test performances are compared using the Wilcoxon rank-sum test to determine statistical significance.

In addition to performance comparison, we conduct a sensitivity analysis across different temperature settings ($\gamma \in {0.5, 0.8, 1.0, 1.2, 1.5}$) to investigate their influence on the exploration–exploitation trade-off. To further understand the model's learning behavior, we analyze the size and structural patterns of the evolved heuristics and examine Transformer training convergence via loss curves. Moreover, to characterize the practical operation of Transformer-guided mutation, we investigate several key aspects: the learned structural regularities, the transparency of the mutation process, the behavior of pattern recombination, and the overall stability of evolution.

Together, these experiments and analyses provide a comprehensive evaluation of TransGP, clarify the conditions under which Transformer-guided mutation offers the greatest advantage, and reveal how the Transformer influences both the efficiency of symbolic search and the structure of the evolved heuristics.

\section{Experimental Results}
\label{sec:results}

\subsection{Compare with GP Methods}

\begin{table}[t]
 \centering
 \footnotesize
 \caption{Mean (std) test performance of the learned heuristics over 30 independent runs across all tasks.}
 \label{tab:final_test_performance}
 \setlength{\tabcolsep}{1mm}
 \begin{threeparttable}
 \begin{tabular}{ccccc}
 \hline
 \textbf{S*} & \textbf{Task} & \textbf{GP} & \textbf{TGP} & \textbf{TransGP} \\
 \hline
 \multirow{3}{*}{1}
  & Task 1 & 1390.99(103.84) & 1321.28(90.85)${(\uparrow)}$ & 1276.79(33.11)${(\uparrow)}$(=) \\
  & Task 2 & 1402.18(79.22) & 1368.10(69.44)(=) & 1342.28(50.07)${(\uparrow)}$(=) \\
  & Task 3 & 1461.44(106.28) & 1484.01(102.21)(=) & 1395.13(64.78)${(\uparrow)(\uparrow)}$ \\
 \hline
 \multirow{3}{*}{2}
  & Task 1 & 580.42(11.17) & 571.84(6.95)${(\uparrow)}$ & 566.21(5.70)${(\uparrow)(\uparrow)}$ \\
  & Task 2 & 605.35(10.50) & 598.08(6.29)${(\uparrow)}$ & 587.20(6.02)${(\uparrow)(\uparrow)}$ \\
  & Task 3 & 627.72(8.16) & 624.28(9.95)(=) & 610.87(4.48)${(\uparrow)(\uparrow)}$ \\
 \hline
 \multirow{3}{*}{3}
  & Task 1 & 248.32(11.82) & 241.02(6.60)${(\uparrow)}$ & 234.87(4.55)${(\uparrow)(\uparrow)}$ \\
  & Task 2 & 270.48(12.74) & 266.10(7.72)(=) & 255.01(4.03)${(\uparrow)(\uparrow)}$ \\
  & Task 3 & 299.04(14.20) & 292.57(7.75)(=) & 278.71(5.57)${(\uparrow)(\uparrow)}$ \\
 \hline
 \end{tabular}
 \begin{tablenotes}
 \scriptsize
 \item S denotes the scenario index; task definitions are consistent with those in Table~\ref{tab:scenarios}. Statistical significance is assessed using the Wilcoxon rank sum test ($p<0.05$). Superscripts indicate the performance of TGP and TransGP compared to the algorithms to their left: ${(\uparrow)}$ significantly better, ${(\downarrow)}$ significantly worse, and (=) not significantly different.
 \end{tablenotes}
 \end{threeparttable}
\end{table}

\begin{figure}[t]
    \centering
    \includegraphics[width=0.99\linewidth]{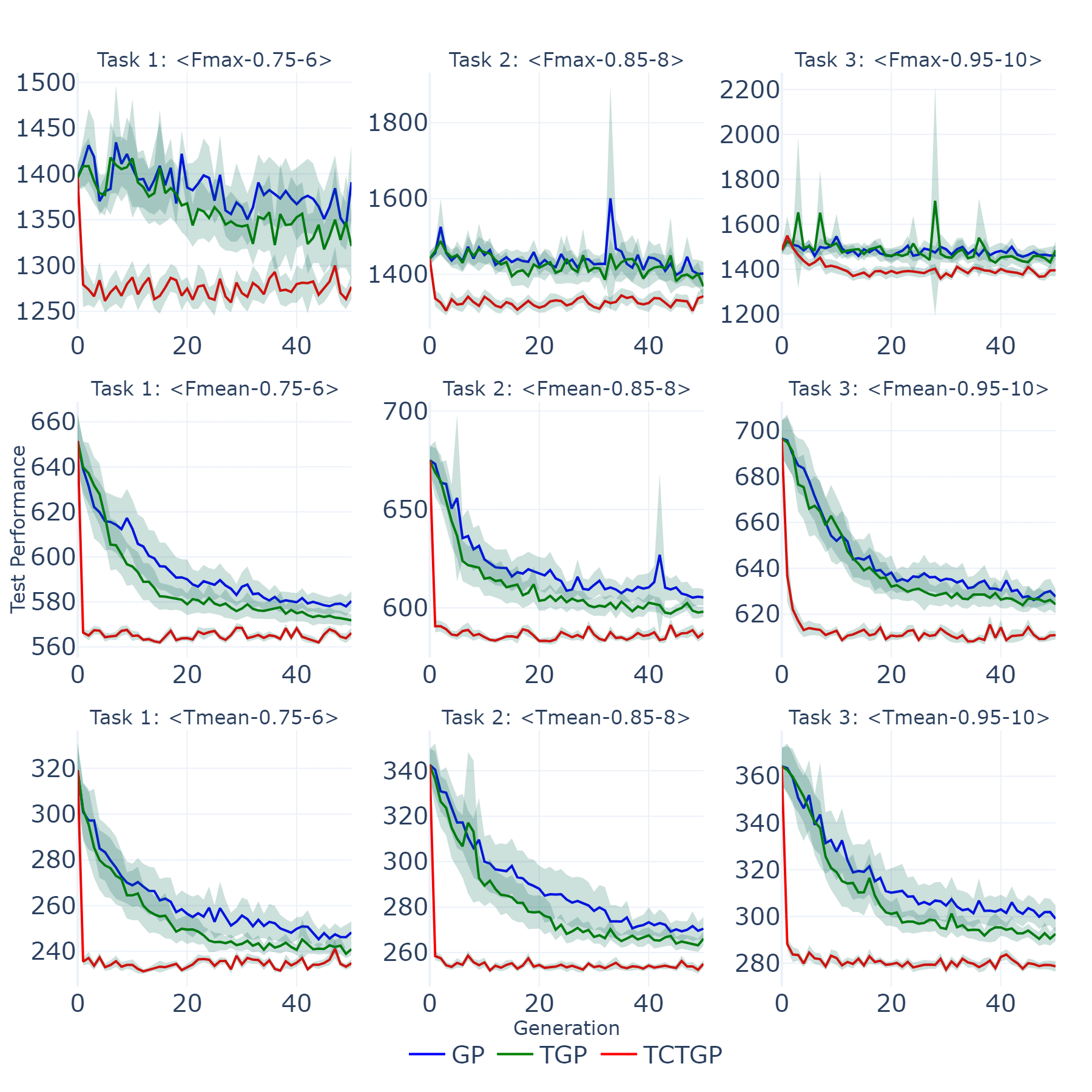}
    \caption{Convergence curves of test performance across 30 independent runs for TransGP and baseline GP methods on three scenarios, each containing three tasks.}
    \label{fig:testperformance}
\end{figure}

\begin{table*}[t]
 \centering
 \footnotesize
 \caption{Mean (std) test performance of the learned heuristics over 30 independent runs of TransGP and handcrafted scheduling rules across all tasks.}
 \label{tab:final_test_performance_handcrafted}
 \setlength{\tabcolsep}{1mm}
 \begin{threeparttable}
 \begin{tabular}{cccccccccccccc}
\hline
\multirow{2}{*}{\textbf{S*}} & \multirow{2}{*}{\textbf{Task}} & \multirow{2}{*}{\textbf{\begin{tabular}[c]{@{}c@{}}SPT\\ NIQ\end{tabular}}} & \multirow{2}{*}{\textbf{\begin{tabular}[c]{@{}c@{}}LPT\\ NIQ\end{tabular}}} & \multirow{2}{*}{\textbf{\begin{tabular}[c]{@{}c@{}}EDD\\ NIQ\end{tabular}}} & \multirow{2}{*}{\textbf{\begin{tabular}[c]{@{}c@{}}FIFO\\ NIQ\end{tabular}}} & \multirow{2}{*}{\textbf{\begin{tabular}[c]{@{}c@{}}SPT\\ WIQ\end{tabular}}} & \multirow{2}{*}{\textbf{\begin{tabular}[c]{@{}c@{}}LPT\\ WIQ\end{tabular}}} & \multirow{2}{*}{\textbf{\begin{tabular}[c]{@{}c@{}}EDD\\ WIQ\end{tabular}}} & \multirow{2}{*}{\textbf{\begin{tabular}[c]{@{}c@{}}FIFO\\ WIQ\end{tabular}}} & \multicolumn{3}{c}{\textbf{PureTrans}} & \multirow{2}{*}{\textbf{TransGP}} \\ \cline{11-13}
                             &                                &                                                                             &                                                                             &                                                                             &                                                                              &                                                                             &                                                                             &                                                                             &                                                                              & min     & mean (std)        & max      &                                   \\ \hline
\multirow{3}{*}{1}           & Task 1                         & 1462.14                                                                     & 1512.95                                                                     & 1517.17                                                                     & 1521.58                                                                      & 1464.61                                                                     & 1518.27                                                                     & 1515.71                                                                     & 1527.31                                                                      & 1257.30 & 2424.80 (2618.89) & 11352.09 & 1276.79(33.11)${(\uparrow)}$      \\
                             & Task 2                         & 1462.11                                                                     & 1563.57                                                                     & 1621.85                                                                     & 1621.09                                                                      & 1473.99                                                                     & 1584.82                                                                     & 1634.58                                                                     & 1615.25                                                                      & 1317.95 & 2448.01 (2546.17) & 12900.12 & 1342.28(50.07)${(\uparrow)}$      \\
                             & Task 3                         & 1589.18                                                                     & 1658.87                                                                     & 1748.73                                                                     & 1731.61                                                                      & 1548.94                                                                     & 1629.52                                                                     & 1717.44                                                                     & 1736.18                                                                      & 1482.35 & 3323.86 (4057.96) & 14999.00 & 1395.13(64.78)${(\uparrow)}$      \\ \hline
\multirow{3}{*}{2}           & Task 1                         & 656.76                                                                      & 680.76                                                                      & 666.13                                                                      & 670.12                                                                       & 655.23                                                                      & 679.21                                                                      & 666.37                                                                      & 668.17                                                                       & 559.97  & 610.00 (96.55)    & 971.03   & 566.21(5.70)${(\uparrow)}$        \\
                             & Task 2                         & 671.27                                                                      & 701.61                                                                      & 687.84                                                                      & 692.09                                                                       & 671.03                                                                      & 704.33                                                                      & 689.16                                                                      & 690.70                                                                       & 579.48  & 880.42 (786.11)   & 3798.87  & 587.20(6.02)${(\uparrow)}$        \\
                             & Task 3                         & 695.21                                                                      & 738.68                                                                      & 720.29                                                                      & 722.00                                                                       & 693.41                                                                      & 736.27                                                                      & 717.41                                                                      & 722.90                                                                       & 682.77  & 1515.65 (1570.67) & 5578.95  & 610.87(4.48)${(\uparrow)}$        \\ \hline
\multirow{3}{*}{3}           & Task 1                         & 324.37                                                                      & 348.26                                                                      & 333.63                                                                      & 337.62                                                                       & 322.84                                                                      & 346.71                                                                      & 333.88                                                                      & 335.67                                                                       & 228.38  & 460.73 (488.58)   & 1987.71  & 234.87(4.55)${(\uparrow)}$        \\
                             & Task 2                         & 338.88                                                                      & 369.13                                                                      & 355.36                                                                      & 359.62                                                                       & 338.64                                                                      & 371.85                                                                      & 356.68                                                                      & 359.22                                                                       & 248.35  & 437.45 (522.61)   & 3014.81  & 255.01(4.03)${(\uparrow)}$        \\
                             & Task 3                         & 362.83                                                                      & 406.20                                                                      & 387.79                                                                      & 389.50                                                                       & 361.03                                                                      & 403.79                                                                      & 384.92                                                                      & 390.41                                                                       & 279.23  & 878.14 (1179.62)  & 4415.67  & 278.71(5.57)${(\uparrow)}$        \\ \hline
\end{tabular}
\begin{tablenotes}
\scriptsize
\item Statistical significance is assessed using the Wilcoxon rank sum test ($p<0.05$) \cite{rosner2003incorporation}. Superscripts indicate the performance of TransGP compared with all handcrafted rules and PureTrans to its left; ${(\uparrow)}$ denotes significantly better performance.
\end{tablenotes}
\end{threeparttable}
\end{table*}

Table~\ref{tab:final_test_performance} reports the mean and standard deviation of test performance for each algorithm, with statistical significance assessed via the Wilcoxon rank sum test ($p < 0.05$) \cite{rosner2003incorporation}. Superscripts indicate the relative performance of TGP and TransGP compared to the algorithms on their left: ($\uparrow$) significantly better, ($\downarrow$) significantly worse, or ($=$) not significantly different. TransGP consistently achieves the best results across all tasks, demonstrating its effectiveness in discovering symbolic scheduling heuristics that minimize both flowtime and tardiness. In scenario 1, TransGP significantly outperforms GP and performs comparably or better than TGP. In scenarios 2 and 3, its advantage becomes more pronounced, showing statistically significant improvements over both baselines on all tasks. The performance gains from GP to TGP highlight the value of task-tagged mutation for cross-task knowledge transfer. TransGP extends this with Transformer-based, task-conditioned mutation, enabling more adaptive and semantic-aware search. Fig.~\ref{fig:testperformance} shows the convergence trends. While all methods improve over generations, TransGP converges faster and more stably, reaching better performance in fewer generations with reduced variance, an important benefit under limited computational budgets. The performance gains from GP to TGP highlight the value of task-tagged mutation for cross-task knowledge transfer. TransGP extends this with Transformer-based, task-conditioned mutation, enabling more adaptive and semantic-aware search. Overall, these findings underscore that combining GP with Transformer-guided, task-conditioned mutation constitutes a non-trivial innovation in symbolic heuristic evolution. It delivers superior performance and effective knowledge transfer across multiple dynamic scheduling tasks.

\subsection{Compare with Handcrafted Rules and Pure Transformer}

To assess the practical effectiveness of TransGP, we compare it against widely used handcrafted scheduling heuristics for DFJSS, as well as against rules generated directly by the trained pure Transformer models (PureTrans). The handcrafted baselines combine classical sequencing rules with standard routing rules commonly employed in real-world dynamic job shop environments:
\begin{itemize}
\item Sequencing rules: SPT (Shortest Processing Time), FIFO (First-In-First-Out), LPT (Longest Processing Time), and EDD (Earliest Due Date);
\item Routing rules: WIQ (Work Remaining in Queue) and NIQ (Number Remaining in Queue).
\end{itemize}

Table~\ref{tab:final_test_performance_handcrafted} reports the mean test performance over 30 independent runs across all scenarios and tasks. The results show that TransGP consistently achieves substantially lower objective values than all handcrafted heuristics, with statistically significant improvements in every task ($p<0.05$). Even the strongest handcrafted combinations, such as SPT+NIQ and SPT+WIQ, remain clearly inferior to TransGP, highlighting the limitations of fixed, single-principle rules in capturing the complex interactions inherent in DFJSS. The results for PureTrans provide additional insights. Although PureTrans is occasionally able to generate high-quality scheduling rules, as reflected by its best-case performance from 30 runs, it exhibits extremely large variance across runs, with mean performance often much worse than TransGP. This instability indicates that, without an evolutionary refinement process, pure Transformer-based rule generation lacks robustness and cannot reliably produce high-quality heuristics.

In contrast, TransGP effectively combines Transformer-generated knowledge with GP, preserving the creative potential of the Transformer while leveraging evolutionary search to filter, refine, and stabilize candidate rules. As a result, TransGP achieves not only superior average performance but also significantly lower variance across runs, demonstrating strong robustness and reliability. Overall, these results confirm that TransGP offers a stable and practically deployable solution, outperforming both handcrafted heuristics and pure Transformer-generated rules in dynamic scheduling environments.

\subsection{Transformer Training Loss}

\begin{figure}[t]
	\centering
	\includegraphics[width=0.98\linewidth]{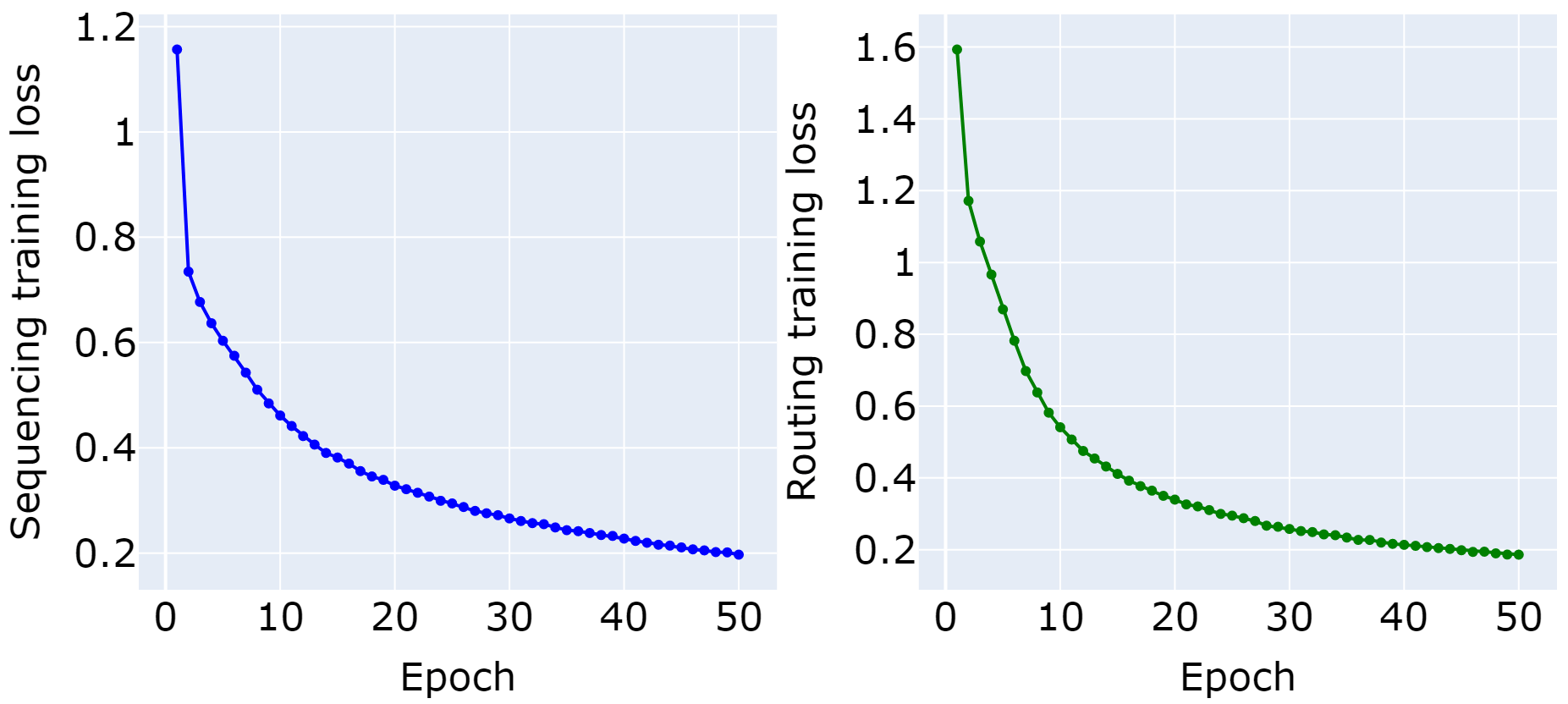}
	\caption{Training loss convergence of the task-conditioned sequencing Transformer and routing Transformer for the \texttt{Fmax} objective.}
	\label{fig:trainingloss}
\end{figure}

Fig.~\ref{fig:trainingloss} presents the training loss curves for both the task-conditioned sequencing Transformer and the routing Transformer when optimizing for the \texttt{Fmax} objective as an example. Both models exhibit smooth and stable convergence without any signs of divergence or loss spikes, indicating that the training process is well-controlled.

Initially, the routing model starts from a higher loss value (1.5932), suggesting greater initial difficulty in learning the routing patterns. In contrast, the sequencing model achieves faster early-stage convergence, reflecting its efficiency in capturing task-specific sequencing behavior. However, by the end of training, both models converged to comparably low loss levels, demonstrating their effectiveness. The absence of instability throughout the training process suggests that the model architectures, optimization strategies, and hyperparameters are well-tuned. These results confirm the robustness of both Transformers in learning their respective symbolic scheduling subcomponents.

\subsection{Temperature Level Impact Analysis}
Temperature $\gamma$ is a crucial hyperparameter in the decoding process of transformer models, especially when generating text. It directly controls the randomness of the model's output. Lower temperature (e.g., $\le$ 1.0) makes the model more deterministic and confident, causing it to pick higher-probability nodes. The output will be more focused, conservative, and often repetitive, but potentially less creative or diverse. Higher temperature (e.g., > 1.0) increases the randomness, allowing the model to choose lower-probability nodes. The output becomes more diverse, creative, and sometimes surprising, but also carries a higher risk of incoherence, nonsensical phrases, or factual errors. This section compares different temperature levels ($\gamma$), exploring the trade-off between output coherence/accuracy and diversity/creativity. 

\begin{table*}[t]
	\centering
	\footnotesize
	\caption{Mean (std) test performance of the learned scheduling heuristics over 30 runs for TransGP under different temperature levels across all tasks and scenarios.}
	\label{tab:test_performance_temp_level}
	\begin{threeparttable}
		\begin{tabular}{ccccccc}
			\hline
			\textbf{S*} & \textbf{Task} & \textbf{$\gamma=0.5$} & \textbf{$\gamma=0.8$} & \textbf{$\gamma=1.0$} & \textbf{$\gamma=1.2$} & \textbf{$\gamma=1.5$} \\
			\hline
			\multirow{3}{*}{1}
			& Task 1 & 1290.74(36.42) & 1275.21(32.03)(=) & 1297.37(31.40)(=)${(\downarrow)}$ & 1276.79(33.11)(=)(=)${(\uparrow)}$ & 1294.58(64.57)(=)(=)(=)(=) \\
			& Task 2 & 1354.26(42.10) & 1361.16(42.97)(=) & 1344.48(48.00)(=)(=) & 1342.28(50.07)(=)(=)(=) & 1347.95(50.01)(=)(=)(=)(=) \\
			& Task 3 & 1394.44(43.08) & 1390.68(60.36)(=) & 1412.18(55.11)(=)(=) & 1395.13(64.78)(=)(=)(=) & 1416.48(68.07)(=)(=)(=)(=) \\
			\hline
			\multirow{3}{*}{2}
			& Task 1 & 564.20(4.10) & 563.55(4.99)(=) & 564.86(4.94)(=)(=) & 566.21(5.70)(=)(=)(=) & 566.99(4.66)${(\downarrow)(\downarrow)}$(=)(=) \\
			& Task 2 & 589.30(5.57) & 586.49(6.09)(=) & 587.10(4.89)(=)(=) & 587.20(6.02)(=)(=)(=) & 587.72(6.26)(=)(=)(=)(=) \\
			& Task 3 & 611.14(6.18) & 610.44(6.69)(=) & 610.71(6.04)(=)(=) & 610.87(4.48)(=)(=)(=) & 614.55(7.98)${(\downarrow)(\downarrow)(\downarrow)(\downarrow)}$ \\
			\hline
			\multirow{3}{*}{3}
			& Task 1 & 232.83(2.44) & 232.94(2.70)(=) & 233.84(3.41)(=)(=) & 234.87(4.55)(=)(=)(=) & 233.49(5.92)(=)(=)(=)(=) \\
			& Task 2 & 253.38(3.46) & 254.60(3.53)(=) & 254.56(4.82)(=)(=) & 255.01(4.03)(=)(=)(=) & 255.54(4.88)(=)(=)(=)(=) \\
			& Task 3 & 276.72(2.94) & 277.74(6.04)(=) & 278.65(4.28)${(\downarrow)}$(=) & 278.71(5.57)${(\downarrow)}$(=)(=) & 281.40(7.83)${(\downarrow)}$(=)(=)(=) \\
			\hline
		\end{tabular}
		\begin{tablenotes}
        \scriptsize
        \item Statistical significance is assessed using the Wilcoxon rank sum test test ($p<0.05$). Superscripts indicate the performance of a given temperature level compared with those to its left: ${(\uparrow)}$ significantly better, ${(\downarrow)}$ significantly worse, and (=) not significantly different.
		\end{tablenotes}
	\end{threeparttable}
\end{table*}

To be specific, TransGP was evaluated using temperature levels $0.5$, $0.8$, $1.0$, $1.2$, and $1.5$. Table~\ref{tab:test_performance_temp_level} reports the mean (std) test performance of the learned scheduling heuristics over 30 runs for TransGP with different temperature levels across all tasks and scenarios. The following observations can be made:
\begin{itemize}
	\item Most pairwise comparisons between temperature levels show no statistically significant difference.
	\item A few comparisons exhibit alternating significance (e.g., in Task 1, $\gamma=1.0$ is significantly worse than $\gamma=0.8$, but $\gamma=1.2$ is significantly better than $\gamma=1.0$), suggesting minor and inconsistent fluctuations.
	\item In Scenario 2, for Task 1 and Task 3, $\gamma=1.5$ shows slightly worse performance than lower temperatures in some cases, but the differences are small and isolated.
\end{itemize}

Overall, although some statistically significant differences appear, particularly at $\gamma=1.5$, the majority of results across all tasks and scenarios indicate no substantial or consistent impact of temperature on performance. Therefore, we conclude that the temperature level does not have a consistently significant effect on TransGP performance. However, very high temperatures such as $\gamma=1.5$ should be used with caution.

\section{Further Analysis: Heuristic Analysis}
\label{analysis}
In this section, we provide an interpretable and systematic analysis of how the proposed TransGP framework optimizes sequencing and routing rules, as well as a detailed examination of the structures, patterns, and cross-task behaviors of the evolved heuristics.

\subsection{Heuristic Size Analysis}
Table~\ref{tab:final_heuristic_size} presents the mean heuristic size (std) of learned scheduling heuristics across 30 runs for each method. Smaller heuristics are preferred for interpretability, particularly in real-world applications where transparency is crucial \cite{xu2025learn}. TransGP consistently generates significantly smaller heuristics in scenarios 1 and 2 across all tasks. In scenario~3, the size advantage is less pronounced, with no significant difference on tasks 2 and 3, suggesting some task dependency.

To better reflect the optimization process, we additionally tracked the size trajectory over generations. Results indicate that TransGP suppresses bloat early in the search, stabilizing around compact, high-performing patterns, in contrast to GP and TGP which continue expanding tree depth without meaningful performance gains. These size reductions stem from the pretrained Transformer's ability to capture generalizable patterns and reusable subtrees across tasks. During mutation, it guides the search toward concise, high-quality structures, avoiding the bloat commonly seen with stochastic operators. This learned guidance minimizes redundancy and promotes efficient symbolic exploration. In contrast, GP and TGP rely on unguided variation, often producing overly complex heuristics with limited performance gains. Overall, TransGP not only improves generalization but also enhances interpretability by evolving more compact and transparent heuristics, underscoring its practical value in real-world scheduling.

\begin{table}[t]
\centering
\footnotesize
\caption{Mean (std) heuristic size of the final generation over 30 runs for each algorithm across tasks.}
\label{tab:final_heuristic_size}
\setlength{\tabcolsep}{1mm}
\begin{tabular}{ccccc}
    \hline
    \textbf{S*} & \textbf{Task} & \textbf{GP} & \textbf{TGP} & \textbf{TransGP} \\
    \hline
    \multirow{3}{*}{1}
    & Task 1 & 80.60(22.08) & 86.20(18.98)(=) & 52.87(16.38)${(\uparrow)(\uparrow)}$ \\
    & Task 2 & 84.13(26.97) & 82.67(20.68)(=) & 66.33(19.94)${(\uparrow)(\uparrow)}$ \\
    & Task 3 & 75.87(21.51) & 84.33(22.04)(=) & 64.20(18.22)(=)${(\uparrow)}$ \\
    \hline
    \multirow{3}{*}{2}
    & Task 1 & 84.87(26.44) & 72.00(22.37)(=) & 57.07(16.70)${(\uparrow)(\uparrow)}$ \\
    & Task 2 & 83.20(20.23) & 73.60(20.14)(=) & 56.67(18.88)${(\uparrow)(\uparrow)}$ \\
    & Task 3 & 77.40(20.81) & 73.67(21.74)(=) & 55.47(16.84)${(\uparrow)(\uparrow)}$ \\
    \hline
    \multirow{3}{*}{3}
    & Task 1 & 82.33(15.91) & 75.80(24.55)(=) & 54.60(23.75)${(\uparrow)(\uparrow)}$ \\
    & Task 2 & 78.60(19.60) & 77.87(23.69)(=) & 88.20(19.26)(=)(=) \\
    & Task 3 & 79.87(21.93) & 71.60(16.50)(=) & 71.40(14.69)(=)(=) \\
    \hline
\end{tabular}
\end{table}

\subsection{Heuristic Structure Analysis Across Tasks}
This section analyzes three sequencing rules and three routing rules drawn from scheduling heuristics learned for three distinct tasks. The tree structures of these rules are shown in Figs.~\ref{fig:tree_structure_sequencing} and \ref{fig:tree_structure_routing}. To provide a comprehensive understanding, we conduct a detailed analysis covering complexity, structural characteristics, function and terminal usage, inter-task rule similarity, and shared subtree patterns. The detailed analysis is presented below.

\subsubsection{Complexity and Structural Characteristics}
\begin{figure}[t]
    \centering
    \includegraphics[width=0.99\linewidth]{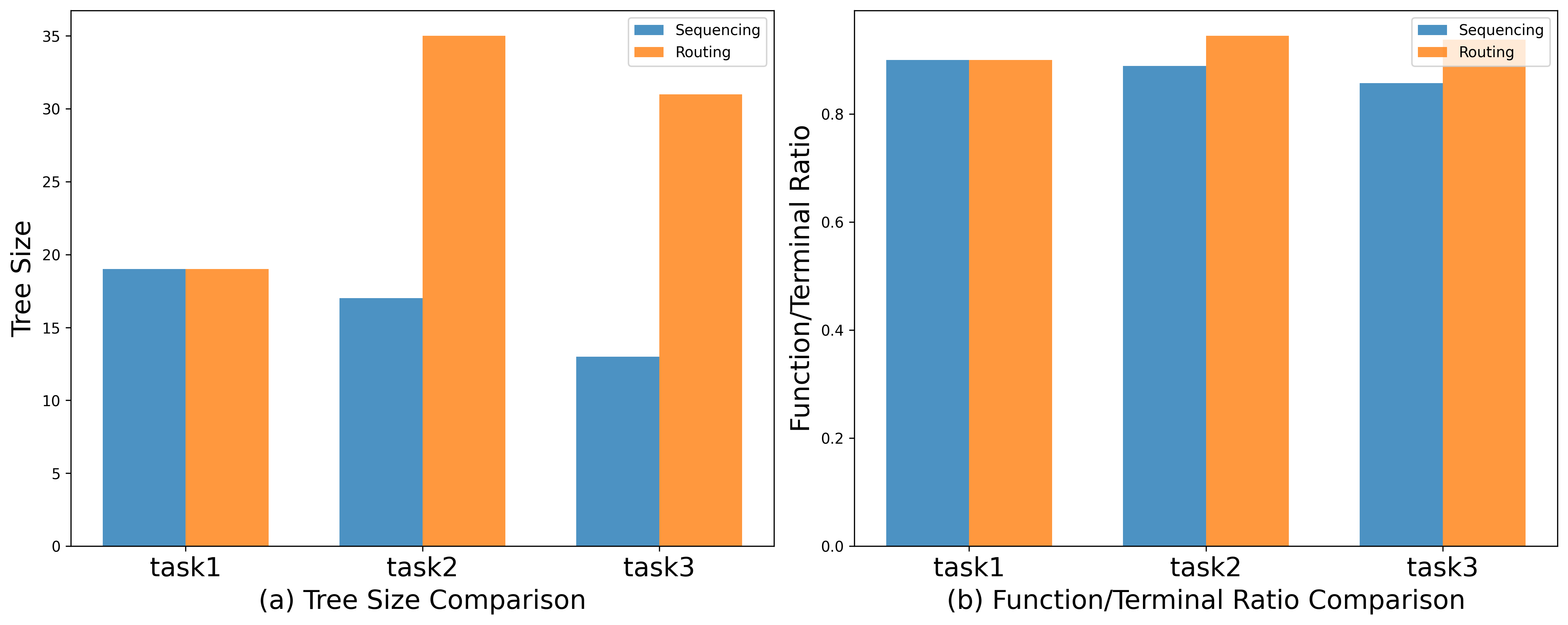}
    \caption{Analysis of rule size and function-to-terminal ratio of the learned heuristics across three tasks from a single run.}
    \label{fig:heuristic_analysis_1}
\end{figure}

The evolved rules exhibit varying complexity by type: sequencing rules are generally more compact, while routing rules are larger and structurally more complex (Fig.~\ref{fig:heuristic_analysis_1}(a)), especially for tasks 2 and 3. The consistently high function-to-terminal ratio (about 0.90–0.94) across all rule types (Fig.~\ref{fig:heuristic_analysis_1}(b)) suggests a strong preference for computational functions over direct terminal usage. This reflects the optimization dynamics of Transformer-guided mutation, which tends to preserve and refine higher-level functional compositions while altering terminals more selectively. By doing so, the model helps encode domain-relevant logic, facilitating the gradual emergence of reusable computational building blocks in both sequencing and routing rules.

\begin{figure}[t]
    \centering
    \includegraphics[width=0.99\linewidth]{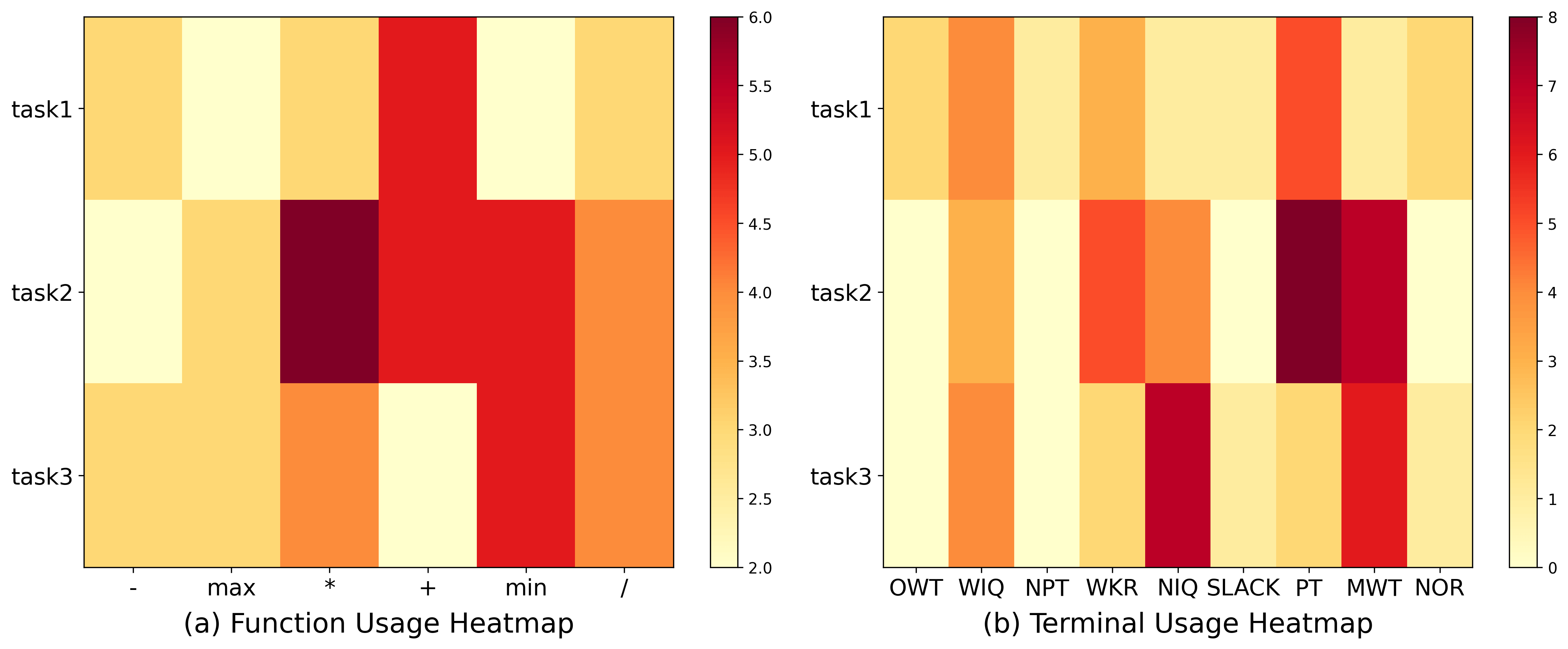}
    \caption{Function and terminal usage heatmap analysis of the learned heuristics across three tasks from a single run.}
    \label{fig:heuristic_analysis_2}
\end{figure}

\subsubsection{Function and Terminal Usage Analysis}
\begin{figure*}[t]
\subfigure[Task 1]{
        \centering
        \includegraphics[width=0.21\textwidth]{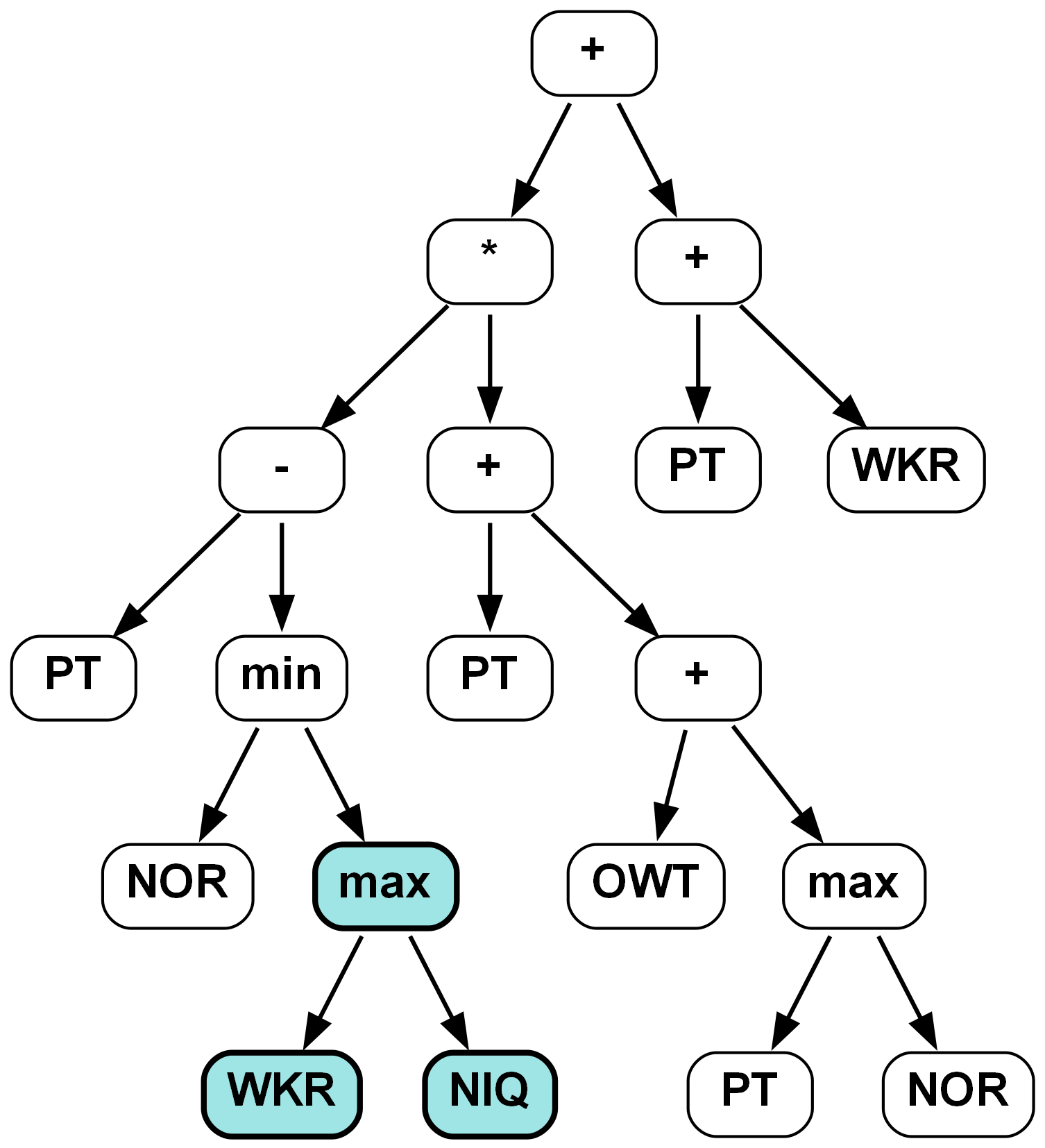}
        \label{fig:tree_sequencing_task1}
}%
\subfigure[Task 2]{
        \centering
        \includegraphics[width=0.24\textwidth]{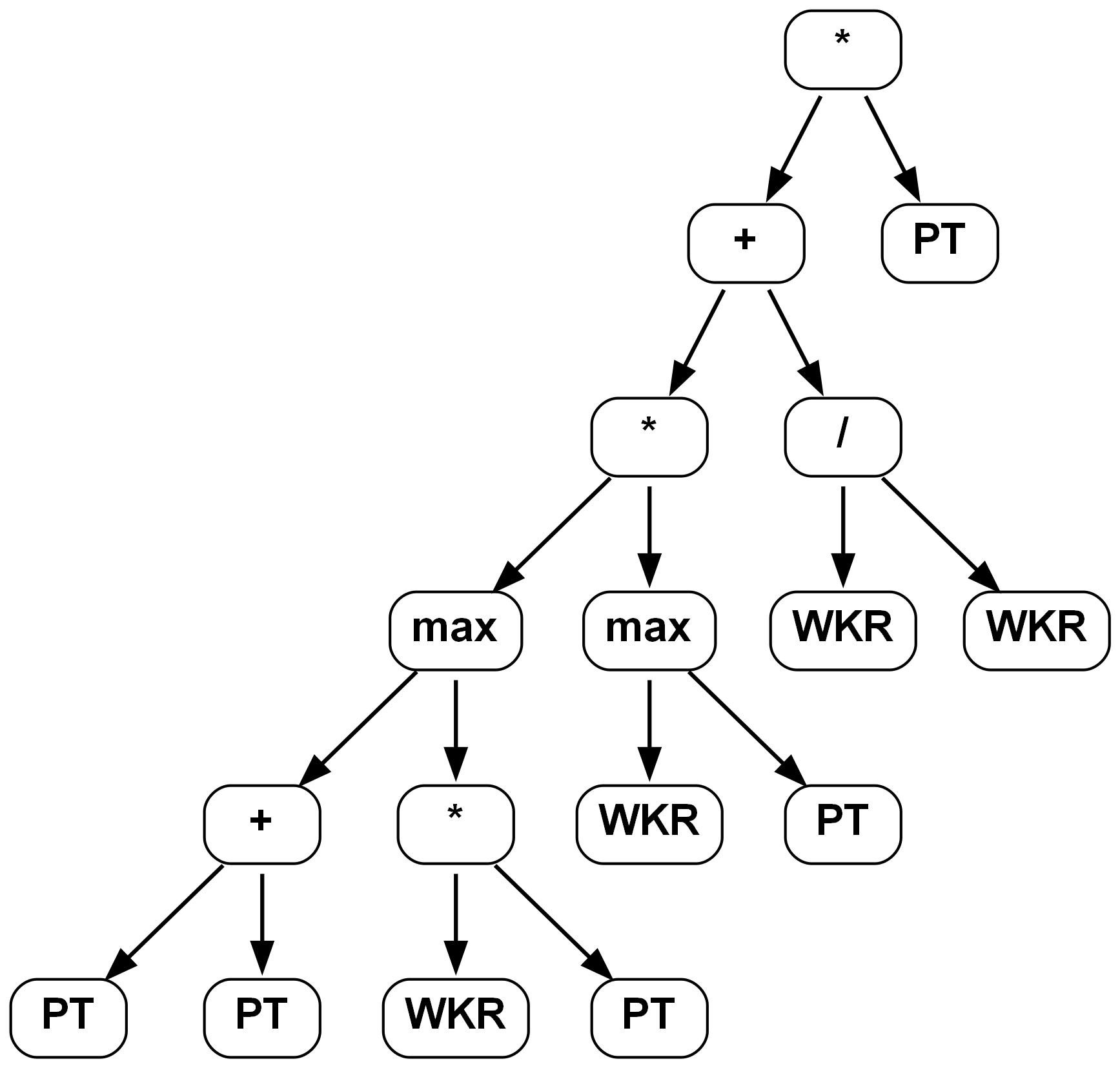}
        \label{fig:tree_sequencing_task2}
}%
\subfigure[Task 3]{
        \centering
        \includegraphics[width=0.16\textwidth]{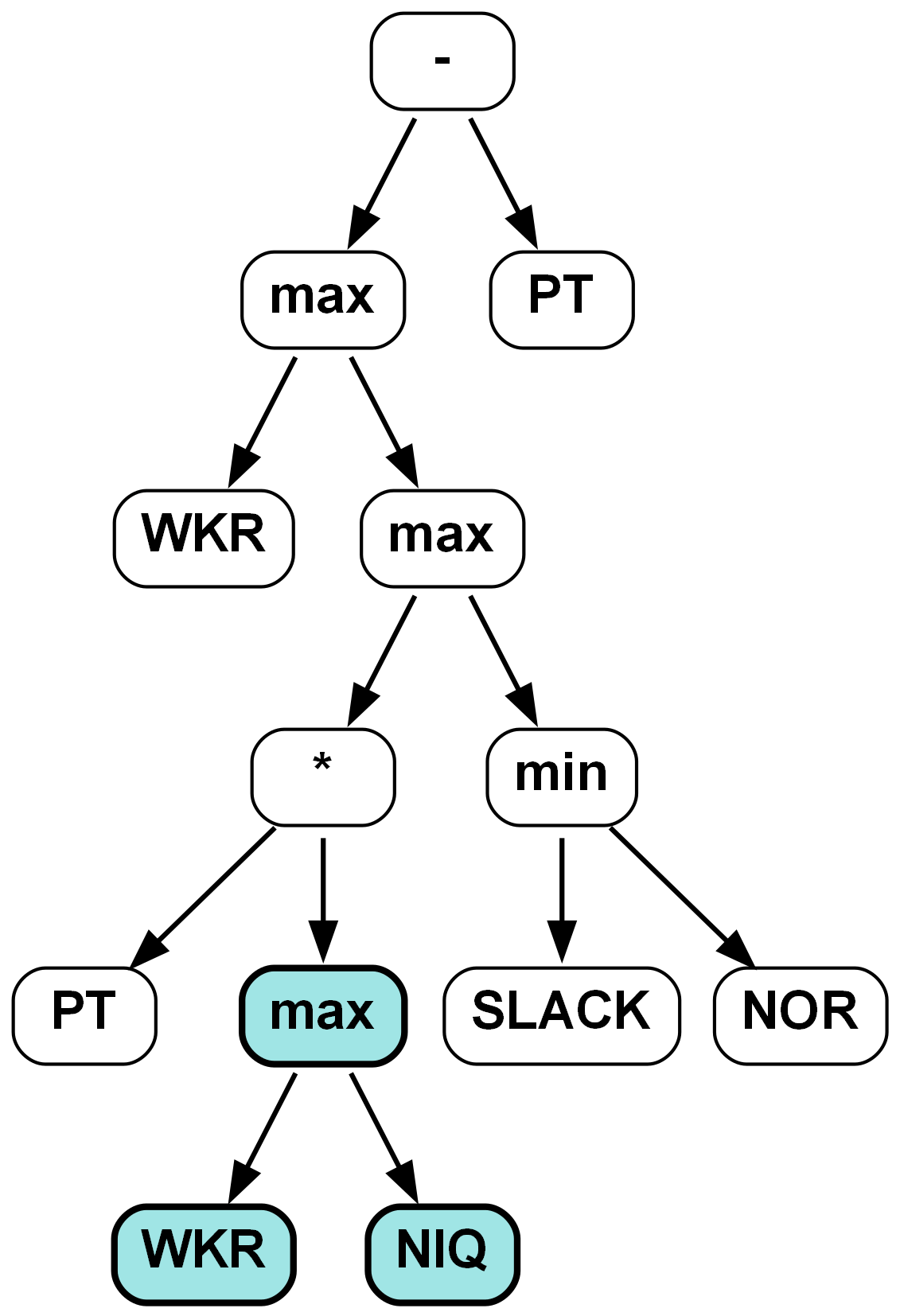}
        \label{fig:tree_sequencing_task3}
}%
\centering
\caption{The tree structures of learned sequencing rules for 3 tasks.}
\label{fig:tree_structure_sequencing}
\end{figure*}

\begin{figure*}[t]
\subfigure[Task 1]{
        \centering
        \includegraphics[width=0.235\textwidth]{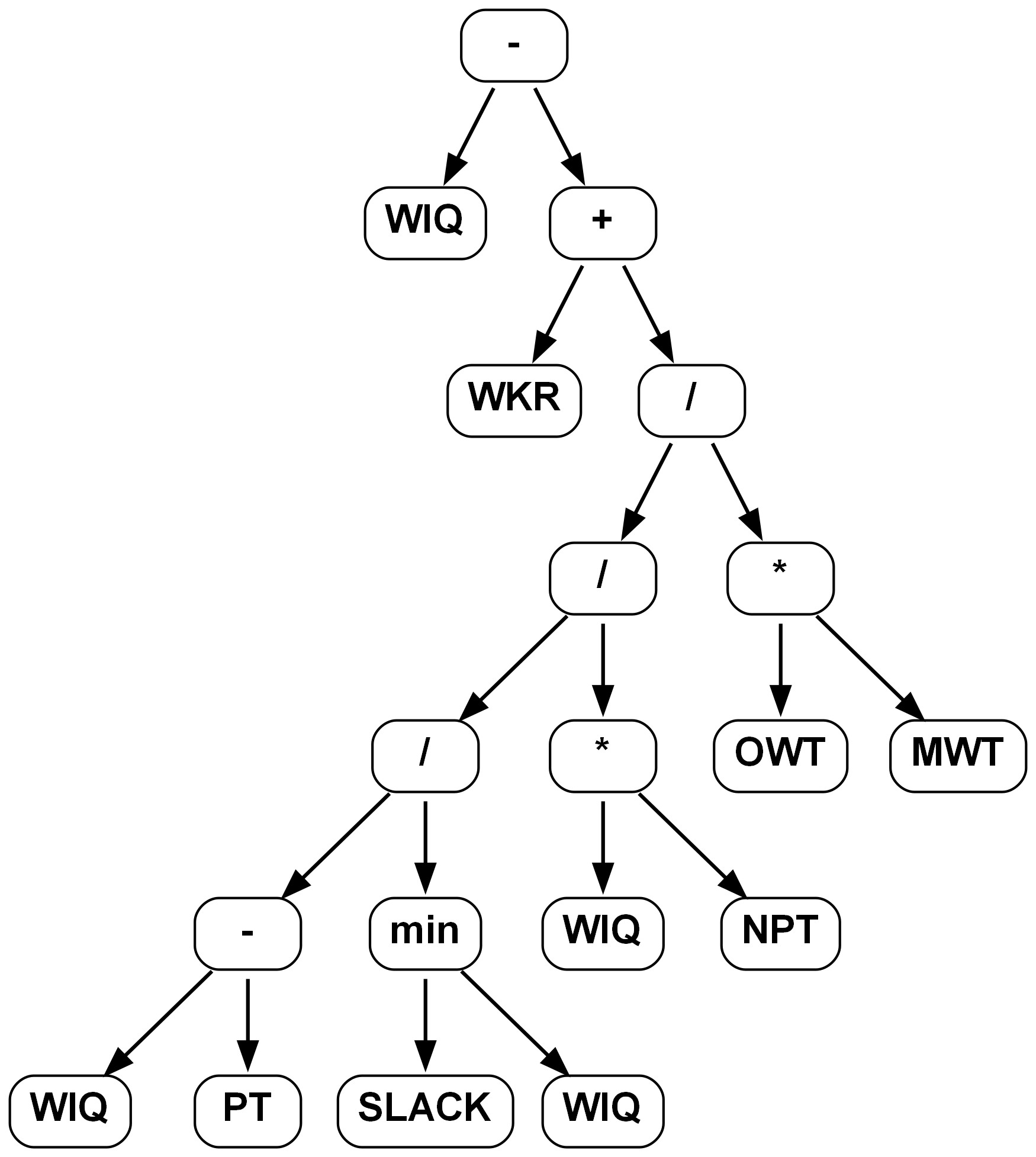}
        \label{fig:tree_routing_task1}
}%
\subfigure[Task 2]{
        \centering
        \includegraphics[width=0.375\textwidth]{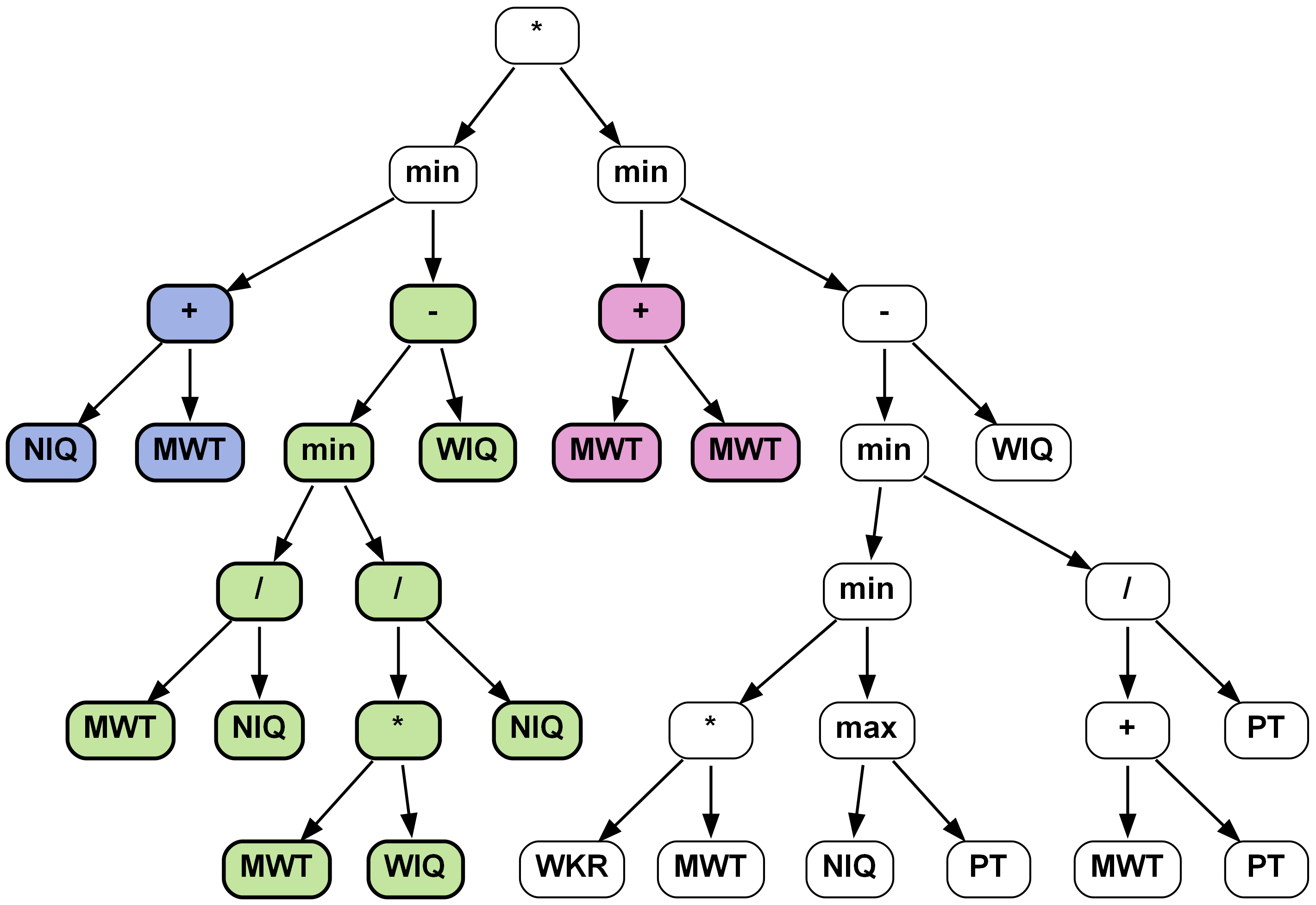}
        \label{fig:tree_routing_task2}
}%
\subfigure[Task 3]{
        \centering
        \includegraphics[width=0.35\textwidth]{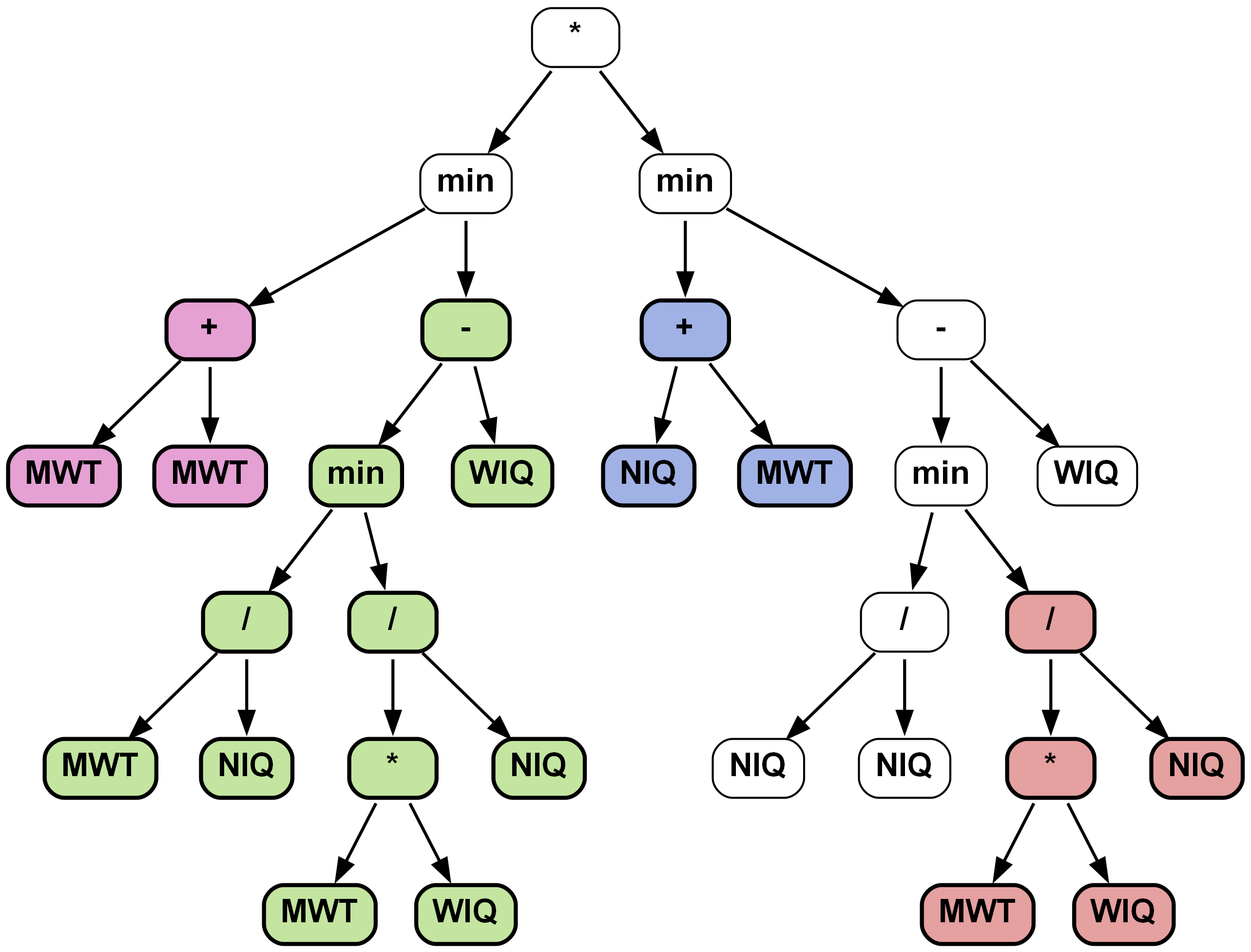}
        \label{fig:tree_routing_task3}
}%
\centering
\caption{The tree structures of learned routing rules for 3 tasks.}
\label{fig:tree_structure_routing}
\end{figure*}

Fig.~\ref{fig:heuristic_analysis_2}(a) shows the distribution of function usage across tasks, with core functions such as \texttt{+}, \texttt{*}, and \texttt{min} frequently employed across all tasks, reflecting their central role in constructing effective decision rules. While many functions are shared, some exhibit task-specific preferences. For instance, \texttt{*} is used notably more often in task 2, indicating adaptation to task characteristics. Fig.~\ref{fig:heuristic_analysis_2}(b) reveals that terminals such as PT, MWT, NIQ, and WKR are broadly used across tasks, though task-specific patterns also emerge: PT appears more frequently in task 2, and NIQ in task 3. Moreover, core terminals like PT and WKR are consistently reused across both sequencing and routing rules within the same task (Figs.~\ref{fig:tree_structure_sequencing} and \ref{fig:tree_structure_routing}), underscoring their general importance in scheduling decisions.

This analysis provides interpretable insight into why certain features dominate: the optimization process tends to preserve terminals that contribute strongly to fitness improvements, forming stable structural anchors around which Transformer-guided mutations build new adaptations.



\subsubsection{Inter-Task Rule Similarity}
To quantify the structural and functional overlap between rules evolved for different tasks, we employ two complementary similarity measures: \emph{Jaccard similarity} and \emph{Size similarity}. These metrics provide insights into how knowledge transfers across tasks and whether TransGP produces task-adapted heuristics while preserving useful structural patterns. 

Jaccard similarity \cite{niwattanakul2013using} quantifies the overlap between two sets by comparing their intersection to their union. For two sets $A$ and $B$, it is defined as:
\begin{equation}
J(A, B) = \frac{|A \cap B|}{|A \cup B|}
\end{equation}
The resulting value ranges from 0 to 1, where $J(A, B) = 1$ indicates that the two sets are identical (sharing all elements), and $J(A, B) = 0$ indicates that they are completely dissimilar (sharing no elements). In this work, we apply Jaccard similarity to compare the terminal sets of evolved heuristics, revealing whether different tasks favor similar or distinct features.

Size similarity quantifies the structural resemblance between two trees by comparing their node counts. For two trees with sizes $\text{size}_1$ and $\text{size}_2$ (where size denotes the number of nodes), it is defined as:
\begin{equation}
\text{Size similarity} = 1 - \frac{|\text{size}_1 - \text{size}_2|}{\max(\text{size}_1, \text{size}_2)}
\end{equation}
The resulting value ranges from 0 to 1, where a value of 1 indicates that the two trees are identical in size, and values approaching 0 indicate substantial size disparity. This metric captures whether structural complexity remains consistent across tasks, which may reflect underlying problem difficulty or evolutionary stability.

These two similarity measures serve distinct but complementary purposes. Jaccard similarity reveals whether heuristics for different tasks rely on shared domain features, indicating potential knowledge transfer or task-specific adaptation. Size similarity, on the other hand, captures whether structural complexity remains consistent across tasks, which may reflect the underlying problem difficulty or the stability of the evolutionary process. Together, these metrics provide a multi-faceted view of how heuristics evolve across heterogeneous tasks and whether TransGP successfully balances task-specific adaptation with structural consistency.

As shown in Table~\ref{tab:similarity_analysis}, sequencing rules for task 1 vs task 3 exhibit high structural similarity (Jaccard: 0.727), while task 2 vs task 3 show lower similarity (Jaccard: 0.364), indicating more divergent rule structures. For routing, task 2 vs task 3 show the highest similarity across both metrics (Jaccard: 0.727, Size: 0.886), suggesting that these tasks share common routing logic and can be addressed using similar heuristics. This cross-task analysis reveals that TransGP captures generalizable structural patterns when task dynamics align, while still allowing divergence when tasks require specialized logic, a key advantage of the task-conditioned generative mutation.

\begin{table}[t]
\centering
\footnotesize
\caption{Similarity analysis of evolved heuristics across tasks.}
\label{tab:similarity_analysis}
\begin{tabular}{cccc}
\hline
\textbf{Comparison} & \textbf{Rule Type} & \textbf{Jaccard} & \textbf{Size} \\
\hline
Task 1 vs Task 2 & sequencing & 0.455 & 0.895 \\
Task 1 vs Task 2 & routing & 0.643 & 0.543 \\
Task 1 vs Task 3 & sequencing & 0.727 & 0.684 \\
Task 1 vs Task 3 & routing & 0.538 & 0.613 \\
Task 2 vs Task 3 & sequencing & 0.364 & 0.765 \\
Task 2 vs Task 3 & routing & 0.727 & 0.886 \\
\hline
\end{tabular}
\end{table}

\subsubsection{Shared Subtree Analysis}
We identify five recurring subtrees across the analyzed rules, as shown in Figs.~\ref{fig:tree_structure_sequencing} and \ref{fig:tree_structure_routing}. This provides strong evidence of modularity and the discovery of reusable computational primitives by the evolutionary process. The most frequently shared pattern, \texttt{/(*(MWT, WIQ), NIQ)}, appears in three locations within the routing rules of task 2 and task 3. The presence of such recurring motifs illustrates that TransGP’s generative operator does not merely search blindly. Rather, it internalizes and reuses effective substructures learned from elite individuals across the population and across tasks. This mechanism directly contributes to both interpretability and improved evolutionary efficiency.

\subsubsection{Summary}
Our analysis demonstrates that the evolutionary process generates heuristics that are not only effective but also structurally interpretable and modular. Rules vary in complexity according to their role, consistently utilize a core set of functions and terminals, and share meaningful substructures across tasks. These findings confirm that TransGP produces transparent and comprehensible routing and sequencing rules, driven by an interpretable and pattern-aware optimization process enabled by the task-conditioned Transformer.

\section{Further Analysis: Understanding Transformer-Guided Mutation}
\label{subsec:transformer_interpretability}
Although the Transformer-guided mutation operator generates symbolic heuristics that are inherently interpretable, understanding how the Transformer internally guides structural modifications is essential for transparency. To this end, we analyze what structural regularities the model learns, how these patterns influence the mutation process, and why specific compositional choices emerge during generation. As a representative case study, we focus on the routing Transformer learned under scenario~2, which consists of task~1: <Fmean-0.75-6>, task~2: <Fmean-0.85-8>, and task~3: <Fmean-0.95-10>.

\begin{table}[t]
\centering
\caption{Top-10 learned structural patterns from elite heuristics across three tasks. Patterns represent common building blocks that the Transformer recombines during mutation.}
\label{tab:learned_patterns}
\footnotesize
\begin{tabular}{c|l|c|c}
\hline
\textbf{Rank} & \textbf{Structural Pattern} & \textbf{Frequency} & \textbf{Coverage (\%)} \\
\hline
1 & \texttt{WIQ} & 1,269 & 5.44 \\
2 & \texttt{*(MWT, TIS)} & 331 & 1.42 \\
3 & \texttt{OWT} & 327 & 1.40 \\
4 & \texttt{*(OWT, PT)} & 295 & 1.26 \\
5 & \texttt{*(WKR, WIQ)} & 280 & 1.20 \\
6 & \texttt{+(NIQ, MWT)} & 264 & 1.13 \\
7 & \texttt{max(min(*(...)} & 253 & 1.08 \\
8 & \texttt{-(min(/(...)} & 247 & 1.06 \\
9 & \texttt{NPT} & 243 & 1.04 \\
10 & \texttt{max(SLACK, MWT)} & 243 & 1.04 \\
\hline
\multicolumn{3}{l|}{\textit{Total coverage of top-10 patterns}} & 16.07 \\
\hline
\end{tabular}
\end{table}

\subsection{Learned Structural Regularities}
\label{subsubsec:learned_patterns}
The Transformer is trained on 3,600 elite routing rules collected across the three tasks, enabling it to model the distribution of high-quality symbolic structures. An analysis of subtree frequencies reveals that the learned distribution is highly concentrated around a limited number of recurring building blocks (subtrees). Table~\ref{tab:learned_patterns} reports the ten most frequent structural patterns extracted from the training set. Simple terminals such as \texttt{WIQ} and \texttt{OWT} appear most frequently, reflecting their broad relevance in routing decisions. More importantly, composite expressions such as \texttt{*(MWT, TIS)} and \texttt{*(WKR, WIQ)} also occur repeatedly, indicating that the Transformer captures meaningful feature interactions rather than isolated terminals. Several nested patterns further indicate that the model incorporates non-trivial hierarchical combinations, such as:


\begin{small}
\begin{verbatim}
  max(min(*(MWT, MWT), PT), +(NIQ, WIQ))
\end{verbatim}
\end{small}

The top-10 patterns collectively account for 16.07\% of all observed subtrees as shown in Table \ref{tab:learned_patterns}, suggesting that high-performing heuristics are built from a compact and reusable set of structural building blocks that the Transformer can subsequently exploit during mutation.

\subsection{Transparency of the Mutation Process}
\label{subsubsec:mutation_transparency}
To illustrate how these learned patterns are applied during evolution, we examine a representative mutation generated from a parent routing rule. The parent rule is given below:
\begin{small}
\begin{verbatim}
  *(min(+(*(WIQ, MWT), MWT), -(min(+(*(WIQ, 
    MWT), MWT), WKR), NIQ)), min(*(WKR, MWT),
      max(/(*(WIQ, WIQ), NIQ), PT)))
\end{verbatim}
\end{small}

\begin{table}[t]
\centering
\caption{Step-by-step token generation for Mutation 1 as shown in Table \ref{tab:mutation_statistics}, showing the Transformer's decision process.}
\label{tab:generation_steps}
\footnotesize
\begin{threeparttable}
\begin{tabular}{c|c|c|c|c|c|c}
\hline
\textbf{Step} & \textbf{Generated} & \textbf{Prob} & \textbf{Top-1} & \textbf{Prob} & \textbf{Top-2} & \textbf{Prob} \\
\hline
1 & \texttt{/} & 0.8547 & \texttt{/} & 0.8547 & \texttt{PT} & 0.0725 \\
2 & \texttt{WKR} & 0.9955 & \texttt{WKR} & 0.9955 & \texttt{+} & 0.0012 \\
3 & \texttt{MWT} & 0.9603 & \texttt{MWT} & 0.9603 & \texttt{WKR} & 0.0255 \\
4 & \texttt{MWT} & 0.9913 & \texttt{MWT} & 0.9913 & \texttt{WKR} & 0.0025 \\
\hline
\end{tabular}
\begin{tablenotes}
\scriptsize
\item The model exhibits high confidence (prob $>0.85$) for 4 out of 4 steps, demonstrating learned compositional preferences.
\end{tablenotes}
\end{threeparttable}
\end{table}

During mutation, the Transformer replaces a subtree at position~8 by generating a new expression token by token. Table~\ref{tab:generation_steps} reports the complete generation trace for Mutation 1 as shown in Table \ref{tab:mutation_statistics}. The model exhibits high confidence in all steps, for instance, assigning a probability of 0.9955 to \texttt{WKR} immediately after selecting \texttt{/}. This indicates that the Transformer has learned strong semantic compatibilities between functions and terminals, rather than sampling randomly. The resulting offspring routing rule is:

\begin{small}
\begin{verbatim}
  *(min(+(*(WIQ, MWT), MWT), /(WKR, MWT)), MWT)
\end{verbatim}
\end{small}

Comparing the parent and offspring shows that the mutation is localized and structurally coherent: the global tree structure is preserved, while a specific subtree is replaced by a composition that frequently appears in the training data. This illustrates how Transformer guidance produces meaningful variations that respect both syntactic validity and learned semantic regularities.

\begin{table}[t]
\centering
\caption{Mutation statistics for five offspring generated from the parent rule.}
\label{tab:mutation_statistics}
\footnotesize
\begin{threeparttable}
\begin{tabular}{l|c}
\hline
\textbf{Metric} & \textbf{Value} \\
\hline
Mutation 1 Similarity (\%) & 100.00  \\
Mutation 2 Similarity (\%) & 57.14  \\
Mutation 3 Similarity (\%) & 100.00  \\
Mutation 4 Similarity (\%) & 57.14  \\
Mutation 5 Similarity (\%) & 57.14  \\ \hline
Average Similarity (\%) & 74.29 ($\pm$21.00) \\
\hline
\end{tabular}
\begin{tablenotes}
\scriptsize
\item Pattern similarity measures the overlap between generated subtrees and training data patterns.
\end{tablenotes}
\end{threeparttable}
\end{table}

\subsection{Pattern Recombination and Stability}
\label{subsubsec:pattern_similarity}
To quantify how consistently learned structures are reused during mutation, we measured the overlap between subtrees in generated offspring and those observed in the training set. Across five offspring generated from the same parent, the average pattern similarity reaches 74.29\% ($\pm$21.00\%), as summarized in Table~\ref{tab:mutation_statistics}. Mutation~1 achieves 100\% similarity, indicating pure recombination of known building blocks, while other offspring introduce partial novelty by combining familiar substructures in new ways. This variability reflects a controlled balance between exploitation and exploration rather than unstable or random generation.

\subsection{Implications for Evolutionary Search}
\label{subsubsec:evolutionary_implications}
Together, these analyses show that Transformer-guided mutation introduces a learned inductive bias into evolutionary search. Instead of relying on random subtree replacement, the Transformer preferentially samples structurally plausible and semantically meaningful expressions drawn from previously successful heuristics. High-confidence predictions reinforce effective building blocks, accelerating convergence, while lower-confidence steps introduce measured diversity that mitigates premature convergence. As a result, mutation remains stochastic but principled, enabling systematic improvement over traditional unguided GP operators.


\section{Conclusion}
\label{conclusion}
This paper proposes TransGP, a hybrid framework that combines task-conditioned generative modeling with GP to evolve interpretable, high-performing scheduling heuristics across multiple tasks. TransGP leverages Transformers trained on successful symbolic heuristics to guide mutation in a semantic-aware, task-adaptive manner. Its key innovation is treating the Transformers as learned mutation operators, enabling targeted exploration of the symbolic search space and effective cross-task knowledge transfer. This addresses the scalability limitations of standard GP and supports adaptive heuristic discovery in dynamic environments. Experiments across diverse scheduling scenarios demonstrate that TransGP consistently outperforms multitask GP baselines, widely used handcrafted heuristics, and pure Transformer. It also improves interpretability by evolving less complex heuristics. Transformer-guided mutation focuses the search on semantically meaningful regions, promoting simpler and more effective structures. Additionally, shared structural patterns across tasks suggest that the learned strategies are both generalizable and reusable.

Future work will explore multi-objective optimization and richer task information to enhance generalization across heterogeneous domains. Overall, TransGP lays the foundation for neuro-symbolic scheduling systems that combine the transparency of symbolic reasoning, the adaptability of generative models, and population-based search capabilities of GP.


%



\section*{Acknowledgment}
The authors gratefully acknowledge the assistance of AI-based language tools, including ChatGPT and Gemini, which were used to refine the writing and enhance the readability of this paper. Typical prompts involved requests such as ``please refine and revise the following content for a paper.''


\ifCLASSOPTIONcaptionsoff
  \newpage
\fi



%



\bibliographystyle{IEEEtran}
\bibliography{reference}

@article{zhang2025llm,
	title={LLM-driven instance-specific heuristic generation and selection},
	author={Zhang, Shaofeng and Liu, Shengcai and Lu, Ning and Wu, Jiahao and Liu, Ji and Ong, Yew-Soon and Tang, Ke},
	journal={arXiv preprint arXiv:2506.00490},
	year={2025}
}

@article{gupta2015multifactorial,
	title={Multifactorial evolution: Toward evolutionary multitasking},
	author={Gupta, Abhishek and Ong, Yew-Soon and Feng, Liang},
	journal={IEEE Transactions on Evolutionary Computation},
	volume={20},
	number={3},
	pages={343--357},
	year={2015},
	publisher={IEEE}
}

@inproceedings{xu2025quality,
	title={Quality Diversity Genetic Programming for Learning Scheduling Heuristics},
	author={Xu, Meng and Neumann, Frank and Neumann, Aneta and Ong, Yew Soon},
	booktitle={Proceedings of the Genetic and Evolutionary Computation Conference},
	pages={1090--1098},
	year={2025}
}

@article{zhong2018multifactorial,
	title={Multifactorial genetic programming for symbolic regression problems},
	author={Zhong, Jinghui and Feng, Liang and Cai, Wentong and Ong, Yew-Soon},
	journal={IEEE transactions on systems, man, and cybernetics: systems},
	volume={50},
	number={11},
	pages={4492--4505},
	year={2018},
	publisher={IEEE}
}

@inproceedings{niwattanakul2013using,
  title={Using of Jaccard coefficient for keywords similarity},
  author={Niwattanakul, Suphakit and Singthongchai, Jatsada and Naenudorn, Ekkachai and Wanapu, Supachanun},
  booktitle={Proceedings of the international multiconference of engineers and computer scientists},
  volume={1},
  number={6},
  pages={380--384},
  year={2013}
}

@article{liu2024evolution,
  title={Evolution of heuristics: Towards efficient automatic algorithm design using large language model},
  author={Liu, Fei and Tong, Xialiang and Yuan, Mingxuan and Lin, Xi and Luo, Fu and Wang, Zhenkun and Lu, Zhichao and Zhang, Qingfu},
  journal={arXiv preprint arXiv:2401.02051},
  year={2024}
}

@article{ye2024reevo,
  title={Reevo: Large language models as hyper-heuristics with reflective evolution},
  author={Ye, Haoran and Wang, Jiarui and Cao, Zhiguang and Berto, Federico and Hua, Chuanbo and Kim, Haeyeon and Park, Jinkyoo and Song, Guojie},
  journal={Advances in neural information processing systems},
  volume={37},
  pages={43571--43608},
  year={2024}
}

@article{novikov2025alphaevolve,
  title={AlphaEvolve: A coding agent for scientific and algorithmic discovery},
  author={Novikov, Alexander and V{\~u}, Ng{\^a}n and Eisenberger, Marvin and Dupont, Emilien and Huang, Po-Sen and Wagner, Adam Zsolt and Shirobokov, Sergey and Kozlovskii, Borislav and Ruiz, Francisco JR and Mehrabian, Abbas and others},
  journal={arXiv preprint arXiv:2506.13131},
  year={2025}
}

@inproceedings{dat2025hsevo,
  title={Hsevo: Elevating automatic heuristic design with diversity-driven harmony search and genetic algorithm using llms},
  author={Dat, Pham Vu Tuan and Doan, Long and Binh, Huynh Thi Thanh},
  booktitle={Proceedings of the AAAI Conference on Artificial Intelligence},
  volume={39},
  number={25},
  pages={26931--26938},
  year={2025}
}

@article{romera2024mathematical,
  title={Mathematical discoveries from program search with large language models},
  author={Romera-Paredes, Bernardino and Barekatain, Mohammadamin and Novikov, Alexander and Balog, Matej and Kumar, M Pawan and Dupont, Emilien and Ruiz, Francisco JR and Ellenberg, Jordan S and Wang, Pengming and Fawzi, Omar and others},
  journal={Nature},
  volume={625},
  number={7995},
  pages={468--475},
  year={2024},
  publisher={Nature Publishing Group UK London}
}

@inproceedings{shady2020automatic,
  title={Automatic design of dispatching rules with genetic programming for dynamic job shop scheduling},
  author={Shady, Salama and Kaihara, Toshiya and Fujii, Nobutada and Kokuryo, Daisuke},
  booktitle={Proceedings of the International Conference on Advances in Production Management Systems},
  pages={399--407},
  year={2020},
  organization={Springer}
}

@article{zeitrag2024cooperative,
  title={A cooperative coevolutionary hyper-heuristic approach to solve lot-sizing and job shop scheduling problems using genetic programming},
  author={Zeitr{\"a}g, Yannik and Rui Figueira, Jos{\'e} and Figueira, Gon{\c{c}}alo},
  journal={International Journal of Production Research},
  volume={62},
  number={16},
  pages={5850--5877},
  year={2024},
  publisher={Taylor \& Francis}
}

@inproceedings{teymourifar2022comparison,
  title={A comparison between linear and non-linear combinations of priority rules for solving flexible job shop scheduling problem},
  author={Teymourifar, Aydin and Li, Jie and Li, Dan and Zheng, Taicheng},
  booktitle={Global Joint Conference on Industrial Engineering and Its Application Areas},
  pages={105--117},
  year={2022},
  organization={Springer}
}

@article{rajendran1999comparative,
  title={A comparative study of dispatching rules in dynamic flowshops and jobshops},
  author={Rajendran, Chandrasekharan and Holthaus, Oliver},
  journal={European journal of operational research},
  volume={116},
  number={1},
  pages={156--170},
  year={1999},
  publisher={Elsevier}
}

@article{zhang2019review,
  title={Review of job shop scheduling research and its new perspectives under Industry 4.0},
  author={Zhang, Jian and Ding, Guofu and Zou, Yisheng and Qin, Shengfeng and Fu, Jianlin},
  journal={Journal of intelligent manufacturing},
  volume={30},
  number={4},
  pages={1809--1830},
  year={2019},
  publisher={Springer}
}

@article{zhang2023survey,
  title={Survey on genetic programming and machine learning techniques for heuristic design in job shop scheduling},
  author={Zhang, Fangfang and Mei, Yi and Nguyen, Su and Zhang, Mengjie},
  journal={IEEE Transactions on Evolutionary Computation},
  volume={28},
  number={1},
  pages={147--167},
  year={2023},
  publisher={IEEE}
}

@article{djurasevic2016adaptive,
  title={Adaptive scheduling on unrelated machines with genetic programming},
  author={{\DJ}urasevi{\'c}, Marko and Jakobovi{\'c}, Domagoj and Kne{\v{z}}evi{\'c}, Karlo},
  journal={Applied Soft Computing},
  volume={48},
  pages={419--430},
  year={2016},
  publisher={Elsevier}
}

@article{xu2025learn,
	title={Learn to optimise for job shop scheduling: a survey with comparison between genetic programming and reinforcement learning},
	author={Xu, Meng and Mei, Yi and Zhang, Fangfang and Zhang, Mengjie},
	journal={Artificial Intelligence Review},
	volume={58},
	number={6},
	pages={1--53},
	year={2025},
	publisher={Springer}
}

@article{vaswani2017attention,
  title={Attention is all you need},
  author={Vaswani, Ashish and Shazeer, Noam and Parmar, Niki and Uszkoreit, Jakob and Jones, Llion and Gomez, Aidan N and Kaiser, {\L}ukasz and Polosukhin, Illia},
  journal={Proceedings of the Advances in Neural Information Processing Systems},
  volume={30},
  year={2017}
}

@inproceedings{hildebrandt2010towards,
  title={Towards improved dispatching rules for complex shop floor scenarios: a genetic programming approach},
  author={Hildebrandt, Torsten and Heger, Jens and Scholz Reiter, Bernd},
  booktitle={Proceedings of the Conference on Genetic and Evolutionary Computation},
  pages={257--264},
  year={2010}
}

@inproceedings{guidotti2024generative,
  title={Generative model for decision trees},
  author={Guidotti, Riccardo and Monreale, Anna and Setzu, Mattia and Volpi, Giulia},
  booktitle={Proceedings of the AAAI Conference on Artificial Intelligence},
  volume={38},
  number={19},
  pages={21116--21124},
  year={2024}
}

@article{lei2023large,
	title={Large-scale dynamic scheduling for flexible job-shop with random arrivals of new jobs by hierarchical reinforcement learning},
	author={Lei, Kun and Guo, Peng and Wang, Yi and Zhang, Jian and Meng, Xiangyin and Qian, Linmao},
	journal={IEEE Transactions on Industrial Informatics},
	volume={20},
	number={1},
	pages={1007--1018},
	year={2023},
	publisher={IEEE}
}

@article{zhou2020automatic,
  title={Automatic design of scheduling policies for dynamic flexible job shop scheduling via surrogate-assisted cooperative co-evolution genetic programming},
  author={Zhou, Yong and Yang, Jianjun and Huang, Zhuang},
  journal={International Journal of Production Research},
  volume={58},
  number={9},
  pages={2561--2580},
  year={2020},
  publisher={Taylor \& Francis}
}

@article{luo2026knowledge,
  title={A Knowledge-Enhanced Evolutionary Multitasking Memetic Algorithm for Multimodal Multiobjective Flexible Job Shop Scheduling Considering Speed},
  author={Luo, Cong and Li, Xinyu and Gao, Liang and Liu, Qihao and Fan, Qingsong},
  journal={IEEE Transactions on Cybernetics},
  year={2026},
  doi={10.1109/TCYB.2026.3662764},
  publisher={IEEE}
}

@article{mei2022explainable,
  title={Explainable artificial intelligence by genetic programming: A survey},
  author={Mei, Yi and Chen, Qi and Lensen, Andrew and Xue, Bing and Zhang, Mengjie},
  journal={IEEE Transactions on Evolutionary Computation},
  volume={27},
  number={3},
  pages={621--641},
  year={2022},
  publisher={IEEE}
}

@article{chen2013flexible,
  title={A flexible dispatching rule for minimizing tardiness in job shop scheduling},
  author={Chen, Binchao and Matis, Timothy I},
  journal={International Journal of Production Economics},
  volume={141},
  number={1},
  pages={360--365},
  year={2013},
  publisher={Elsevier}
}

@article{rosner2003incorporation,
  title={Incorporation of clustering effects for the Wilcoxon rank sum test: a large-sample approach},
  author={Rosner, Bernard and Glynn, Robert J and Ting Lee, Mei-Ling},
  journal={Biometrics},
  volume={59},
  number={4},
  pages={1089--1098},
  year={2003},
  publisher={Oxford University Press}
}

@article{guo2024improved,
  title={An improved genetic programming hyper-heuristic for the dynamic flexible job shop scheduling problem with reconfigurable manufacturing cells},
  author={Guo, Haoxin and Liu, Jianhua and Wang, Yue and Zhuang, Cunbo},
  journal={Journal of Manufacturing Systems},
  volume={74},
  pages={252--263},
  year={2024},
  publisher={Elsevier}
}

@article{shady2022novel,
  title={A novel feature selection for evolving compact dispatching rules using genetic programming for dynamic job shop scheduling},
  author={Shady, Salama and Kaihara, Toshiya and Fujii, Nobutada and Kokuryo, Daisuke},
  journal={International Journal of Production Research},
  volume={60},
  number={13},
  pages={4025--4048},
  year={2022},
  publisher={Taylor \& Francis}
}

@article{durasevic2023heuristic,
  title={Heuristic and metaheuristic methods for the parallel unrelated machines scheduling problem: a survey},
  author={{\DJ}urasevi{\'c}, Marko and Jakobovi{\'c}, Domagoj},
  journal={Artificial Intelligence Review},
  volume={56},
  number={4},
  pages={3181--3289},
  year={2023},
  publisher={Springer}
}

@inproceedings{chen2024generate,
  title={Generate a Single Heuristic for Multiple Dynamic Flexible Job Shop Scheduling Tasks by Genetic Programming},
  author={Chen, Jiayin and Jia, Yahui and Bi, Ying and Chen, Weineng},
  booktitle={Proceedings of the IEEE Congress on Evolutionary Computation},
  pages={1--8},
  year={2024},
  organization={IEEE}
}

@article{braune2022genetic,
  title={A genetic programming learning approach to generate dispatching rules for flexible shop scheduling problems},
  author={Braune, Roland and Benda, Frank and Doerner, Karl F and Hartl, Richard F},
  journal={International Journal of Production Economics},
  volume={243},
  pages={108342},
  year={2022},
  publisher={Elsevier}
}

@article{zhang2022task,
  title={Task relatedness-based multitask genetic programming for dynamic flexible job shop scheduling},
  author={Zhang, Fangfang and Mei, Yi and Nguyen, Su and Tan, Kay Chen and Zhang, Mengjie},
  journal={IEEE Transactions on Evolutionary Computation},
  volume={27},
  number={6},
  pages={1705--1719},
  year={2022},
  publisher={IEEE}
}

@article{zhang2021surrogate,
  title={Surrogate-assisted multitask genetic programming for learning scheduling heuristics},
  author={Zhang, Fangfang and Nguyen, Su and Mei, Yi and Zhang, Mengjie},
  journal={Genetic Programming for Production Scheduling: An Evolutionary Learning Approach},
  pages={291--311},
  year={2021},
  publisher={Springer}
}

@article{chen2025optimizing,
  title={Optimizing dynamic flexible job shop scheduling using an evolutionary multi-task optimization framework and genetic programming},
  author={Chen, Xiaolong and Li, Junqing and Wang, Zunxun and Chen, Qingda and Gao, Kaizhou and Pan, Quanke},
  journal={IEEE Transactions on Evolutionary Computation},
  year={2025},
  doi={10.1109/TEVC.2025.3543770},
  publisher={IEEE}
}

@article{yi2025improved,
  title={An improved deep Q-network for dynamic flexible job shop scheduling with limited maintenance resources},
  author={Yi, Wenchao and Chen, Nanxing and Chen, Yong and Pei, Zhi},
  journal={International Journal of Production Research},
  volume={63},
  number={23},
  pages={9112--9133},
  year={2025},
  publisher={Taylor \& Francis}
}

@article{xu2023genetic,
  title={Genetic programming for dynamic flexible job shop scheduling: Evolution with single individuals and ensembles},
  author={Xu, Meng and Mei, Yi and Zhang, Fangfang and Zhang, Mengjie},
  journal={IEEE Transactions on Evolutionary Computation},
  volume={28},
  number={6},
  pages={1761--1775},
  year={2023},
  publisher={IEEE}
}

@article{wang2021bi,
  title={A bi-level framework for learning to solve combinatorial optimization on graphs},
  author={Wang, Runzhong and Hua, Zhigang and Liu, Gan and Zhang, Jiayi and Yan, Junchi and Qi, Feng and Yang, Shuang and Zhou, Jun and Yang, Xiaokang},
  journal={Proceedings of the Advances in Neural Information Processing Systems},
  volume={34},
  pages={21453--21466},
  year={2021}
}

@article{corsini2024self,
  title={Self-labeling the job shop scheduling problem},
  author={Corsini, Andrea and Porrello, Angelo and Calderara, Simone and Dell'Amico, Mauro},
  journal={Proceedings of the Advances in Neural Information Processing Systems},
  volume={37},
  pages={105528--105551},
  year={2024}
}

@article{zhang2020learning,
  title={Learning to dispatch for job shop scheduling via deep reinforcement learning},
  author={Zhang, Cong and Song, Wen and Cao, Zhiguang and Zhang, Jie and Tan, Puay Siew and Chi, Xu},
  journal={Proceedings of the Advances in Neural Information Processing Dystems},
  volume={33},
  pages={1621--1632},
  year={2020}
}

@inproceedings{kotary2022fast,
  title={Fast approximations for job shop scheduling: A lagrangian dual deep learning method},
  author={Kotary, James and Fioretto, Ferdinando and Van Hentenryck, Pascal},
  booktitle={Proceedings of the AAAI Conference on Artificial Intelligence},
  volume={36},
  number={7},
  pages={7239--7246},
  year={2022}
}

@inproceedings{ding2019two,
  title={A two-individual based evolutionary algorithm for the flexible job shop scheduling problem},
  author={Ding, Junwen and L{\"u}, Zhipeng and Li, Chu-Min and Shen, Liji and Xu, Liping and Glover, Fred},
  booktitle={Proceedings of the AAAI conference on Artificial Intelligence},
  volume={33},
  number={01},
  pages={2262--2271},
  year={2019}
}

@inproceedings{zhang2024deep,
title={Deep Reinforcement Learning Guided Improvement Heuristic for Job Shop Scheduling},
author={Zhang, Cong and Cao, Zhiguang and Song, Wen and Wu, Yaoxin and Zhang, Jie},
booktitle={Proceedings of the International Conference on Learning Representations},
year={2024}
}

@article{chen2025genetic,
  title={Genetic Programming with Reinforcement Learning Trained Transformer for Real-World Dynamic Scheduling Problems},
  author={Chen, Xian and Qu, Rong and Dong, Jing and Bai, Ruibin and Jin, Yaochu},
  journal={arXiv preprint arXiv:2504.07779},
  year={2025}
}

@inproceedings{tao2023program,
  title={Program synthesis with generative pre-trained transformers and grammar-guided genetic programming grammar},
  author={Tao, Ning and Ventresque, Anthony and Saber, Takfarinas},
  booktitle={Proceedings of the IEEE Latin American Conference on Computational Intelligence},
  pages={1--6},
  year={2023},
  organization={IEEE}
}

@article{gontier2020measuring,
  title={Measuring systematic generalization in neural proof generation with transformers},
  author={Gontier, Nicolas and Sinha, Koustuv and Reddy, Siva and Pal, Chris},
  journal={Proceedings of the Advances in Neural Information Processing Systems},
  volume={33},
  pages={22231--22242},
  year={2020}
}

@article{nguyen2012computational,
  title={A computational study of representations in genetic programming to evolve dispatching rules for the job shop scheduling problem},
  author={Nguyen, Su and Zhang, Mengjie and Johnston, Mark and Tan, Kay Chen},
  journal={IEEE Transactions on Evolutionary Computation},
  volume={17},
  number={5},
  pages={621--639},
  year={2012},
  publisher={IEEE}
}

@book{koza1999genetic,
  title={Genetic programming III: Darwinian invention and problem solving},
  author={Koza, John R},
  volume={3},
  year={1999},
  publisher={Morgan Kaufmann}
}

@article{bagal2021molgpt,
  title={MolGPT: molecular generation using a transformer-decoder model},
  author={Bagal, Viraj and Aggarwal, Rishal and Vinod, PK and Priyakumar, U Deva},
  journal={Journal of chemical information and modeling},
  volume={62},
  number={9},
  pages={2064--2076},
  year={2021},
  publisher={ACS Publications}
}

@article{han2021transformer,
  title={Transformer in transformer},
  author={Han, Kai and Xiao, An and Wu, Enhua and Guo, Jianyuan and Xu, Chunjing and Wang, Yunhe},
  journal={Proceedings of the Advances in Neural Information Processing Systems},
  volume={34},
  pages={15908--15919},
  year={2021}
}

\end{document}